%% file: neurips_2025.tex
\documentclass{article}

\usepackage[preprint,nonatbib]{neurips_2025}

\usepackage[utf8]{inputenc} 
\usepackage[T1]{fontenc}    
\usepackage[square,numbers]{natbib}
\usepackage[pagebackref=true,breaklinks=true,colorlinks,citecolor={green},urlcolor={purple}]{hyperref}
\usepackage{url}            
\usepackage{booktabs}       
\usepackage[table]{xcolor} 
\usepackage{amsfonts}       
\usepackage{nicefrac}       
\usepackage{microtype}      
\usepackage{xcolor}         

\usepackage{multicol}
\usepackage{multirow}

\usepackage{bbding}
\usepackage{pifont}
\usepackage{tabularx}
\usepackage{graphicx}
\usepackage{subfigure}
\usepackage{amsmath}
\usepackage{cancel}
\usepackage{caption}
\usepackage{enumitem}
\usepackage[capitalize]{cleveref}

\definecolor{mycyan}{cmyk}{.1,0,0,0}
\definecolor{mygray}{gray}{.95}
\definecolor{mypink}{rgb}{.99,.91,.95}

\newcommand{\cmark}{\ding{51}}%
\newcommand{\cmarkg}{\textcolor{lightgray}{\ding{51}}}%
\newcommand{\xmark}{\ding{55}}%
\newcommand{\xmarkg}{\textcolor{lightgray}{\ding{55}}}%

\newcommand{\bluecref}[1]{%
    \textcolor{blue}{{\hypersetup{linkcolor=blue}%
     \edef\temp{\noexpand\cref{#1}}\temp}}%
}
\newcommand{\blueref}[1]{%
    \textcolor{blue}{{\hypersetup{linkcolor=blue}%
     \edef\temp{\noexpand\ref{#1}}\temp}}%
}

\title{Does Your 3D Encoder Really Work? When Pretrain-SFT from 2D VLMs Meets 3D VLMs}

\author{
    Haoyuan Li$^{1*}$, Yanpeng Zhou$^{2}$, Yufei Gao$^{1}$, Tao Tang$^{1}$, 
    Jianhua Han$^{2}$, Yujie Yuan$^{2}$,\\[5pt]
    \textbf{Dave Zhenyu Chen}$^{2}$, \textbf{Jiawang Bian}$^{3}$, \textbf{Hang Xu}$^{2}$, \textbf{Xiaodan Liang}$^{1,4,5\dag}$\\[5pt]
    $^1$Shenzhen campus of Sun Yat-sen University, 
    $^2$Huawei Noah's Ark Lab, 
    $^3$MBZUAI,\\[5pt]
    $^4$Peng Cheng Laboratory, 
    $^5$Guangdong Key Laboratory of Big Data Analysis and Processing\\ [5pt]
    \url{https://github.com/Li-Hao-yuan/3DRDQA}
}

\begin{document}

\maketitle

\begin{center}
    \centering
    \captionsetup{type=figure}
    \begin{center}
        \includegraphics[width=1\textwidth]{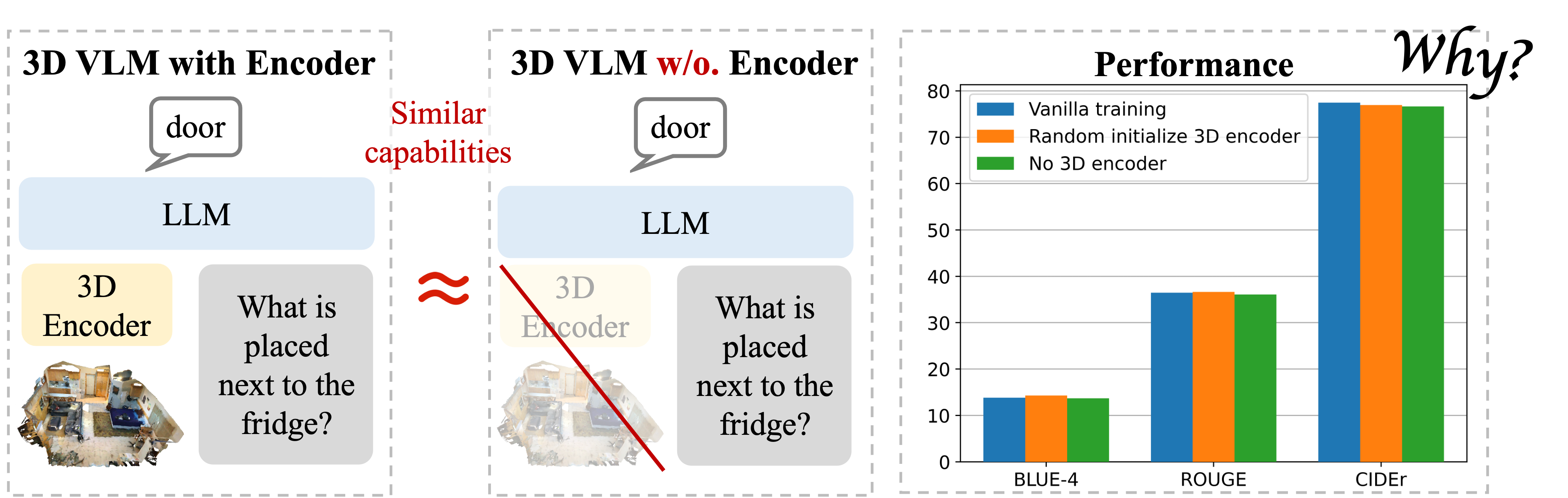}
    \end{center}
 \caption{\textbf{Left:} 3D VLM (Vision Language Model) with encoder leverages 3D Encoder to "see" scenes for question answering. \textbf{Middle:} 3D VLM without Encoder direct outputs answer. \textbf{Right:} 3D VLMs with and without an encoder achieve similar performance, but why?}
    \label{fig:teaser}
    \vspace{-3mm}
\end{center}

\renewcommand{\thefootnote}{$\ast$}
\footnotetext{Work done as an intern at Huawei Noah's Ark Lab.}

\renewcommand{\thefootnote}{$\dag$}
\footnotetext{Corresponding author.}

\input{sec/0_abstract}
\input{sec/1_intro}

\input{sec/2_relatedwork}

\input{sec/3_motivation}

\input{sec/4_method}
\input{sec/5_conclusion}

{
    \small
    \bibliographystyle{ieeenat_fullname}
    \bibliography{main}
}

\input{sec/X_suppl}

\end{document}

%% file: sec/0_abstract.tex
\begin{abstract}

Remarkable progress in 2D Vision-Language Models (VLMs) has spurred interest in extending them to 3D settings for tasks like 3D Question Answering, Dense Captioning, and Visual Grounding. Unlike 2D VLMs that typically process images through an image encoder, 3D scenes, with their intricate spatial structures, allow for diverse model architectures. Based on their encoder design, this paper categorizes recent 3D VLMs into 3D object-centric, 2D image-based, and 3D scene-centric approaches. Despite the architectural similarity of 3D scene-centric VLMs to their 2D counterparts, they have exhibited comparatively lower performance compared with the latest 3D object-centric and 2D image-based approaches. To understand this gap, we conduct an in-depth analysis, revealing that 3D scene-centric VLMs show limited reliance on the 3D scene encoder, and the pre-train stage appears less effective than in 2D VLMs. Furthermore, we observe that data scaling benefits are less pronounced on larger datasets. Our investigation suggests that while these models possess cross-modal alignment capabilities, they tend to over-rely on linguistic cues and overfit to frequent answer distributions, thereby diminishing the effective utilization of the 3D encoder. To address these limitations and encourage genuine 3D scene understanding, we introduce a novel 3D Relevance Discrimination QA dataset designed to disrupt shortcut learning and improve 3D understanding. Our findings highlight the need for advanced evaluation and improved strategies for better 3D understanding in 3D VLMs.

\end{abstract}

%% file: sec/1_intro.tex
\section{Introduction}
\label{sec:intro}

The remarkable progress of 2D Vision-Language Models (VLMs) through pre-training and supervised fine-tuning (SFT)~\cite{li2023blip,mishra2024image,chen2024internvl,dong2024internlm,liu2023visual,zhu2023minigpt,wu2024visionllm,wang2024qwen2,team2023gemini,liu2024deepseek} has sparked increasing interest in extending these models to 3D settings~\cite{hong20233d,wang2023chat,huang2023chat,li20243dmit,huang2023embodied,zhu2024llava,chen2024ll3da,chen2024grounded,zhi2024lscenellm}. By leveraging powerful open-source Large Language Models (LLMs) and richly annotated 3D datasets~\cite{dai2017scannet,azuma2022scanqa,chen2020scanrefer,achlioptas2020referit3d,ma2022sqa3d,chen2021scan2cap}, substantial progress has been made in 3D Vision-Language tasks such as \textbf{3D} \textbf{Q}uestion \textbf{A}nswer (3D-QA)~\cite{azuma2022scanqa,ma2022sqa3d}, \textbf{3D} \textbf{D}ense \textbf{C}aptioning (3D-DC) ~\cite{chen2024vote2cap,chen2021scan2cap} and \textbf{3D} \textbf{V}isual \textbf{G}rounding (3D-VG)~\cite{chen2020scanrefer,achlioptas2020referit3d}.

Unlike the common practice in 2D VLMs of typically utilizing image encoders, 3D scenes, as complex spatial structures comprising various object relationships, can be approached as combinations of different modalities, leading to diverse model design patterns. 
As shown in \cref{fig:related_work}, based on the encoder employed, recent works can be categorized into three main types: i) \textbf{3D object-centric VLM}, which understand space as a collection of objects and model individual objects and their relationships; ii) \textbf{2D image-based VLM}, which interpret space as a continuous video sequence and derive spatial understanding from video analysis; and iii) \textbf{3D scene-centric VLM}, which treat each scene as a holistic entity and directly reason about the scene itself.

Leveraging advancements in modality alignment for 3D object encoders and 2D image encoders through contrastive learning, both 3D object-centric and 2D image-based VLMs have significantly surpassed 3D scene-centric VLMs in performance. Despite 3D scene-centric VLMs exhibiting the most similar model design to 2D VLMs, it has not emerged as the most prevalent or successful approach in the field. 
We analyze this performance gap by comparing 3D scene-centric approaches with successful experience from 2D VLMs and begin with three key observations:

\begin{figure*}[t]
    \centerline{\includegraphics[width=\textwidth]{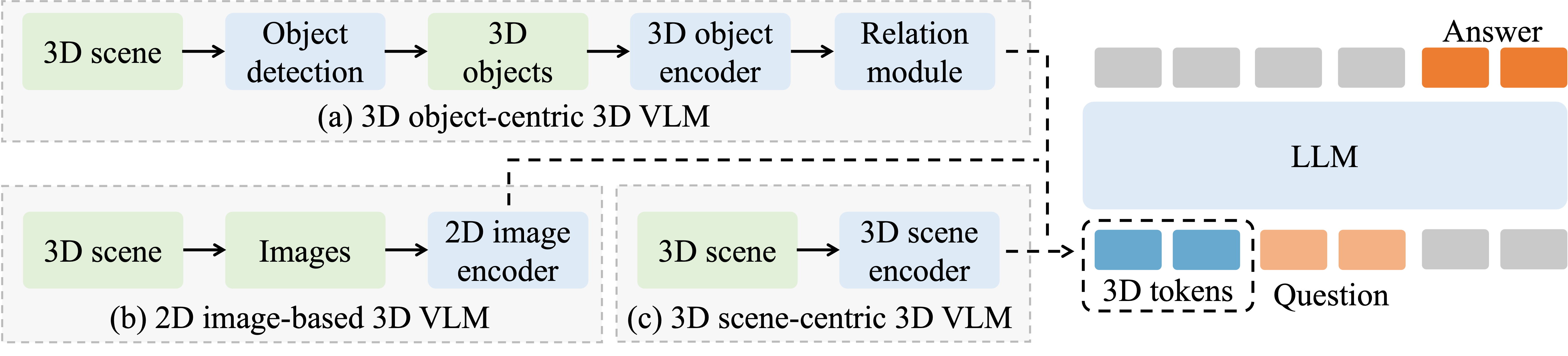}}
    \caption{\textbf{Visualization of different 3D VLM patterns.}
    Similar to 2D VLM, 3D VLM also requires an encoder to extract features that serve as 3D tokens for the cross-modal input.
    Variations in the model design primarily stem from the choice of encoder: (a) utilizing a 3D object encoder necessitates initial object detection and subsequent relation modeling, (b) employing a 2D image encoder requires rendering the 3D scene into a sequence of images, and (c) directly processing the 3D scene.
    }
    \vspace{-3mm}
    \label{fig:related_work}
\end{figure*}

\vspace{-1mm}
\begin{itemize}[leftmargin=20pt]
    \item \textbf{Observation 1:} \emph{3D scene-centric VLMs achieve comparable performance even without the 3D scene encoder's pre-trained weights.}
    \item \textbf{Observation 2:} \emph{In contrast to 2D VLMs, the pre-training stage appears to have a less significant effect on 3D scene-centric VLMs.}
    \item \textbf{Observation 3:} \emph{3D scene-centric VLMs exhibit data scaling when trained on small-scale datasets, but not on large-scale datasets.}
\end{itemize}
\vspace{-1mm}

To better understand and address these unexpected phenomena, we first use CLIP to encode scene descriptions, obtaining text tokens known to be well-aligned. Compared to leveraging the 3D tokens extracted from 3D encoder under the same settings, we find that the 3D scene-centric VLM does not lack the ability to align 3D tokens with text. We then focus on the question format in 3D-QA. By designing a multiple-choice version of ScanQA, named ScanQA-Choice, we demonstrate that 3D scene-centric VLMs tend to over-rely on textual information, making them not directly adaptable to multiple-choice formats. Subsequently, we analyze the distribution of the model's generated answers under different evaluation settings, revealing a significant overfitting to the most frequent answer distributions in the 3D datasets, thus negating the necessity of utilizing the 3D encoder. Finally, based on the above findings, we create "poisoned" copies of the data in ScanQA-Choice where the 3D tokens are manipulated. These poisoned samples, along with the original data, form the 3D Relevance Discrimination QA (3D-RDQA) pair dataset. 
The 3D-RDQA dataset, designed to disrupt the reliance on learning superficial question-answer relationships and encourage 3D scene understanding, enables subsequent experiments to further validate our findings.
To summarize, our key contributions lie in:
\begin{itemize}[leftmargin=20pt]
    \item We first quantitatively analyze the 3D scene-centric VLMs' reliance on 3D geometry. Experiments show a limited capacity to leverage the 3D spatial structure effectively, which may consequently diminish the importance of the learned 3D scene encoder.
    \item We design a multiple-choice 3D QA task demonstrating the over-reliance on language over 3D reasoning and identify overfitting to frequent answers as a key reason for the limited utility of the 3D scene encoder.
    \item We introduce a novel 3D Relevance Discrimination QA dataset to break shortcut learning and promote genuine 3D scene understanding.    
\end{itemize}

%% file: sec/2_relatedwork.tex
\vspace{-5mm}
\section{Related Work}
\label{sec:related_work}
\vspace{-2mm}
\subsection{3D-vision Large Language Models}
\vspace{-2mm}

The rapid advancement of pre-trained LLMs and their demonstrated strong comprehension and reasoning capabilities have significantly promoted the considerable progress of 3D VLMs. 
Researchers initially leverage off-the-shelf 3D Encoder~\cite{Uni3d,yu2022point,zhang2023learning,zhang2022pointclip,zhu2023pointclip,huang2023clip2point,zeng2023clip2,xue2023ulip,xue2024ulip,li2025unigs,liu2023openshape} pretrained on large-scale text-image-3D triplets for 3D object understanding~\cite{tang2025more,yu2022point,tang2024minigpt,guo2023point}, while latest advancement has demonstrated improved performance by embedding the 3D encoder within the LLM itself~\cite{tang2025exploring}.
However, the inherent spatial complexity of 3D scenes, encompassing a richer array of objects and intricate distance relationships, presents a significant challenge for the direct transferability of conventional contrastive learning paradigms to 3D scene encoders. Furthermore, the scarcity of large-scale datasets comprising aligned text-image-3D scene data has resulted in the absence of available pre-trained encoders with rich semantic information specifically for 3D scenes. Based on the 3D encoder employed, existing methodologies can be broadly categorized into three distinct groups:

\textbf{3D object-centric 3D VLM.} Object-centric approaches view spatial understanding as dealing with a collection of objects, which enables reusing existing 3D object encoders. These methods usually start by finding all the individual objects in a scene using instance segmentation or object detection. Then, they utilize an available 3D object encoder to get semantic information and a relationship module to model how these objects are spatially related. 
LEO~\cite{huang2023embodied} adopts PointNet++~\cite{qi2017pointnet++} to encode 3D object features and Spatial Transformer~\cite{chen2022language} for modeling point cloud embedding of all objects into object-centric 3D token embeddings. Chat-3D and Chat-3D v2~\cite{wang2023chat} leverage off-the-shelf 3D segmentation models~\cite {jiang2020pointgroup,misra2021end,qi2019deep} for instance segmentation, which is later encoded and modeled through 3D object encoder and relation module to extract scene features. 3DMiT~\cite{li20243dmit} utilize parallel 3D scene and object encoder~\cite{huang2022frozen,xue2024ulip,Uni3d} for global scene and local object visual features.

\textbf{2D image-based 3D VLM.} Image-based methods, on the other hand, treat spatial understanding like a video taken in a space. This means they can easily connect to existing 2D VLM research as a specific type of multi-view images understanding task. With powerful 2D image encoders, these methods often perform better than those using direct 3D input. 
LLaVA-3D~\cite{zhu2024llava} combines monocular depth and camera pose to obtain spatial position embedding for multi-view image tokens for overall scene understanding.

\textbf{3D scene-centric 3D VLM.} With the lack of available 3D scene encoders with rich semantic information, 3D scene-centric methods directly use 3D object detection or 3D scene segmentation models as 3D scene encoders to obtain spatial information. 
3D-LLM~\cite{hong20233d} introduces a family of LLM-driven 3D generalist models capable of processing a wide range of textual instructions using 3D features reconstructed from multi-view images. LL3DA~\cite{chen2024ll3da} leverages Vote2Cap-DETR~\cite{chen2023end,chen2024vote2cap} to extract scene features and object proposals for the object-centric task. Grounded 3D-LLM~\cite{chen2024grounded} proposes Contrastive Language-Scene Pre-training to pre-train a 3D point cloud encoder and a cross-modal interactor for multi-task instruction tuning. LSceneLLM~\cite{zhi2024lscenellm} focuses on fine-grained understanding and proposes an adaptive self-attention module and dense vision token selector to dynamically sample question-related tokens.

While 3D object-centric and 2D image-based approaches offer promising ways for tackling 3D scene understanding and have demonstrated encouraging results, we believe that scene-centric methods also hold significant potential for advancement. Consequently, our research will primarily focus on exploring and developing 3D scene-centric methods. We aim to investigate how direct processing of the entire 3D scene, without explicit object decomposition or reliance on 2D projections, can lead to robust and comprehensive spatial understanding. This direction warrants further exploration to fully realize its capabilities and address the inherent challenges associated with directly encoding complex 3D environments.

\subsection{Training stages of 3D VLMs}
The pre-train and SFT two-stage training has been shown to work well for 2D VLMs. In the pre-training step, only the projector between the model and the encoder is trained for better alignment. Then, SFT uses higher-quality and efficient data for instruction tuning. 
Following this idea, \cite{zhu2024llava,wang2023chat,huang2023chat,huang2023embodied,hong20233d,chen2024grounded} train the model with pre-train alignment and SFT tuning. In contrast, LL3DA~\cite{chen2024ll3da} only trains the Q-Former~\cite{li2023blip} for connecting the 3D encoder and LLM, while ~\cite{zhi2024lscenellm,li20243dmit} directly fine-tune the projector and LLMs. 
However, upon closer examination of existing 3D scene-centric approaches, we observe notable variations in the training paradigms employed by models in LL3DA~\cite{chen2024ll3da}, LSceneLLM~\cite{zhi2024lscenellm}, 3D-LLM~\cite{hong20233d}, and Grounded 3D-LLM~\cite{chen2024grounded}. This divergence in training strategies suggests a lack of unified understanding or consensus within the research community regarding the optimal training methodology for scene-centric 3D scene understanding with LLM, which highlights the need for further investigation into effective and consistent training protocols for this promising direction.

%% file: sec/3_motivation.tex
\section{Problem Analysis}
\label{sec:motivation}

\begin{table*}[h]
	\centering
  \addtolength{\tabcolsep}{5pt}
  \setlength{\tabcolsep}{3pt}
 \begin{tabular}{ c | c c c | c c c }
\toprule
LLM  & Encoder weight & Encoder output & Q-Former output & BLUE-4 $\uparrow$ & CIDEr$\uparrow$ & ROUGE $\uparrow$ \\
 \midrule  
    \multicolumn{3}{l}{\textit{\textbf{Official LL3DA}}} \\
    \midrule
 Opt-1.3B & \cmarkg & \cmarkg & \cmarkg & 13.53 & 76.69 & 37.31 \\
 
 \midrule 
    \multicolumn{3}{l}{\textit{\textbf{Encoder ablation}}} \\
    \midrule
    
\multirow{5}{*}{Qwen2-1.5B}  & \cmarkg & \cmarkg & \cmarkg & 13.82 & 77.44 & 36.48 \\
  & \colorbox{mypink}\xmark & \cmarkg & \cmarkg & 14.31 & 76.96 & 36.63 \\
 & \xmarkg & \colorbox{mypink}\xmark & \cmarkg & 13.68 & 76.64 & 36.09 \\
 & \xmarkg & \xmarkg & \colorbox{mypink}\xmark & 0.00 & 1.12 & 3.67 \\

\bottomrule

\end{tabular}
      \caption{\textbf{Analysis of 3D tokens utilization.} Under the same settings as LL3DA, we conduct further analysis to investigate the impact of randomly initialized encoder weights, utilizing encoder outputs, and employing Q-Former outputs. Results demonstrate that LL3DA understands different 3D scenes with the same query from Q-Former.}
      \label{tab:LL3DA_encoder_ablation}
    \vspace{-3mm}
\end{table*}

\textbf{Our work focuses on the in-depth analysis of 3D scene-centric approaches. For brevity, from now on, we will use 3D VLM and 3D Encoder to denote 3D scene-centric VLM and 3D scene encoder, respectively.}
To facilitate a more thorough analysis of the impact of 3D encoders and variations in training stages, we select ScanQA as our primary benchmark and adopt LL3DA as our baseline model. LL3DA's relatively simple architecture and its demonstrated strong performance on 3D-QA and 3D-DC make it a suitable starting point for our investigations.

\subsection{Does your pre-trained 3D encoder work?}
\label{sec:motivaiton_sub1}

Following the setup of LL3DA, the Q-Former outputs only 32 tokens. Increasing the number of input 3D tokens did not lead to significant improvements, suggesting limited utilization of the input 3D tokens. As demonstrated in \cref{tab:LL3DA_encoder_ablation}, our subsequent ablation experiments reveal that the understanding of 3D scene information heavily relies on the scene-agnostic latent queries learned by the Q-Former, rather than the features extracted by the 3D scene encoder itself. Consequently, when we do not load the 3D scene encoder pre-trained weights or zero out all features extracted by it, the model's baseline performance remains largely unaffected. This finding emphasizes a potential inefficiency in how 3D VLM integrates and utilizes 3D tokens.

\subsection{Dose pre-training stage matter?}
\label{sec:motivaiton_sub2}

\begin{table*}[h]
	\centering
 \begin{tabular}{ c | c | c c | c c c }
\toprule
LLM & Pre-train & SFT & Encoder weight & BLUE-4 $\uparrow$ & CIDEr$\uparrow$ & ROUGE $\uparrow$ \\

 \midrule 
    \multicolumn{3}{l}{\textit{\textbf{Official LL3DA}}} \\
    \midrule
 Opt-1.3B & \cmarkg & \xmarkg & \cmarkg & 13.53 & 76.69 & 37.31 \\

 \midrule 
    \multicolumn{3}{l}{\textit{\textbf{Pre-train stage ablation}}} \\
    \midrule

\multirow{2}{*}{Qwen2-1.5B} & \multirow{2}{*}{\xmark} & \colorbox{mycyan}\cmark & \xmarkg & 10.84 & 71.22 & 37.43 \\
& & \cmarkg & \colorbox{mycyan}\cmark & 13.33 & 77.23 & 37.10  \\

\midrule

\multirow{2}{*}{Qwen2-1.5B} & \multirow{2}{*}{\cmark} & \colorbox{mycyan}\cmark & \xmarkg & 10.88\textcolor{blue}{(+0.04)} & 70.40\textcolor{orange}{(-0.82)} & 36.80\textcolor{orange}{(-0.63)} \\
 & & \cmarkg & \colorbox{mycyan}\cmark & 14.58\textcolor{blue}{(+1.25)} & 77.03\textcolor{orange}{(-0.20)} & 37.80\textcolor{blue}{(+0.70)}  \\

\bottomrule

\end{tabular}
      \caption{\textbf{Analysis of pre-train stage.} Further analysis of SFT stage and Encoder weight. (*) denotes performance change compared to no pre-train stage.}
      \label{tab:LL3DA_pretrain_ablation}
    \vspace{-6mm}
\end{table*}

In the training paradigm of 2D VLMs, the pre-train stage typically involves alignment using a broad range of less refined data, which facilitates subsequent SFT. To replicate this setup in 3D VLMs, 3D VLMs should similarly pre-train on object-agnostic data such as scene descriptions, followed by SFT on object-centric datasets like ScanQA.
However, we observe an anomalous loss trend during the training stage transition. More specifically, when entering the SFT stage, the loss begins to converge from a very high initial value, similar to the loss observed when starting directly with SFT. This suggests that the pre-training process does not provide a substantial benefit to the final SFT performance on ScanQA. 

Following the training paradigm of \cite{liu2023visual}, we perform one epoch each of pre-training and SFT on the same dataset. As shown in \cref{tab:LL3DA_pretrain_ablation}, a comparison of performance with and without the pre-train stage reveals no significant improvement. Furthermore, we observed performance variations depending on whether we randomly initialize encoder weights. Compared with results in \cref{tab:LL3DA_encoder_ablation}, this suggests an increased utilization of the 3D encoder during the SFT stage. However, the model still achieved considerable performance without pre-trained weights.

\subsection{Does 3D VLMs have scaling capabilities?}

\begin{table*}[h]
	\centering
  \addtolength{\tabcolsep}{5pt}
  \setlength{\tabcolsep}{3pt}
 \begin{tabular}{ c | c c c c | c c c }
\toprule
\multirow{2}{*}{LLM} & \multicolumn{2}{c}{QA} & \multicolumn{2}{c|}{Densecap} & \multirow{2}{*}{BLUE-4 $\uparrow$} & \multirow{2}{*}{CIDEr$\uparrow$} & \multirow{2}{*}{ROUGE $\uparrow$} \\

  & ScanQA & 3D-LLM QA & ScanRefer & Nr3D & & & \\
 
\midrule  
    \multicolumn{3}{l}{\textit{\textbf{Data scaling}}} \\
    \midrule

\multirow{4}{*}{Qwen2-1.5B} & \colorbox{mycyan}{\cmark} & \xmarkg & \xmarkg & \xmarkg & 11.12 & 70.06 & 36.47 \\
& \cmarkg & \colorbox{mycyan}{\cmark} & \xmarkg & \xmarkg & 10.90 & 70.88 & 36.00 \\
 & \cmarkg & \cmarkg & \colorbox{mycyan}{\cmark} & \xmarkg & 12.46 & 73.49 & 36.82 \\
 & \cmarkg & \cmarkg & \cmarkg & \colorbox{mycyan}\cmark & 14.31 & 76.96 & 36.63 \\
 \midrule  
    \multicolumn{3}{l}{\textit{\textbf{Model scaling}}} \\
    \midrule
 
Qwen2-1.5B & \cmarkg & \cmarkg & \cmarkg & \cmarkg & 14.31 & 76.96 & 36.63 \\
Qwen2-7B  & \cmarkg & \cmarkg & \cmarkg & \cmarkg & 13.67 & 81.43 & 38.37 \\

\bottomrule

\end{tabular}
      \caption{\textbf{Analysis of scaling capabilities.} We take ScanQA as the benchmark for evaluation of scaling capabilities under pre-train and SFT stages.}
      \label{tab:scaling_baltion}
\end{table*}

\begin{table*}[h]
	\centering
  \setlength{\tabcolsep}{4pt}
 \begin{tabular}{ c | c | c c c | l}
\toprule
LLM & Dataset & BLUE-4 $\uparrow$ & CIDEr$\uparrow$ & ROUGE $\uparrow$ & Update \\

\midrule  

\multirow{5}{*}{Qwen2-1.5B} & 145k & 13.33 & 77.23 & 37.10 & Same setting with LL3DA~\cite{chen2024ll3da}\\
 & 162k & 13.63 & 77.38 & 36.80 & further +3D-LLM QA~\cite{huang2023embodied}\\
 & 263k & 12.87 & 76.42 & 36.56 & further +Multi3DRefer~\cite{zhang2023multi3drefer}$\&$Scan2Cap~\cite{chen2021scan2cap}\\
 & 355k & 13.64 & 78.56 & 37.44 & further +SQA3D~\cite{ma2022sqa3d}$\&$3RScanQA~\cite{wald2019rio}\\
 & 661k & 12.65 & 77.31 & 37.43 & further +Scene Alignment from  
\cite{huang2023embodied}\\
\midrule
Qwen2-7B & 661k & 14.43 & 81.52 & 38.57 & further use lager LLM\\
\bottomrule


\end{tabular}
      \caption{\textbf{Analysis of large-scale scaling capabilities.} Large-scale dataset scaling capabilities with pre-train and SFT stages.}
      \label{tab:further_data_scaling_ablation_scanqa}
    \vspace{-3mm}
\end{table*}

We further analyze the scaling capabilities of 3D VLM on ScanQA, and the analysis on 3D-DC can be found in \textbf{\textcolor{blue}{Supplementary Material \blueref{sec:appendix_data_scaling_densecap}}}. As shown in \cref{tab:scaling_baltion}, progressively increasing the data scale leads to a corresponding gradual improvement in performance. Similarly, switching to larger models results in improvement in CIDEr and ROUGE scores. However, incorporating 3D-LLM QA does not enhance performance, while improvements are only observed after scaling up the 3D-DC dataset. 
Therefore, we further scale up the data in \cref{tab:further_data_scaling_ablation_scanqa}. The results indicate that 3D VLM no longer exhibits a significant data scaling capability when scaling up data size over 135k. While model scaling remains effective for larger datasets, further increasing the data size does not yield significant performance gains for larger LLMs.

To summarize, our findings indicate that 3D VLMs demonstrate model scaling potential, although a considerable performance gap remains compared to current leading approaches~\cite{zhu2024llava,huang2023embodied}. Moreover, the capacity of data scaling is only evident on small-scale datasets with cross-task data and does not scale effectively to larger datasets.

%% file: sec/4_method.tex
\section{Method \& Experiments}

Following the three observations in \cref{sec:motivation}, we will investigate the potential impact of three key aspects on 3D VLMs: the lack of semantic information in the 3D Encoder, the question-answering format within 3D VLM, and the distribution of data used for training. We will explore these directions to better understand their influence on the overall performance and capabilities of 3D VLMs.

\subsection{Ablation of semantic information}

\begin{table*}[h]
	\centering
  \addtolength{\tabcolsep}{5pt}
  \setlength{\tabcolsep}{3pt}
 \begin{tabular}{  c c | c c c }
\toprule
Multi-modal input & Pre-train  & BLUE-4 $\uparrow$ & CIDEr$\uparrow$ & ROUGE $\uparrow$ \\
\midrule

 \multirow{2}{*}{\text{Scene description}} & ScanQA* & 5.18 & 72.42 & 26.68 \\
  & 3D-LLM Pre & 5.40 & 75.74 & 27.78 \\

\midrule

\multirow{2}{*}{\text{3D scene}} & ScanQA* & 5.37 & 77.55 & 28.35 \\
  & 3D-LLM Pre & 5.54 & 76.30 & 27.92 \\

\bottomrule

\end{tabular}
      \caption{\textbf{Analysis of semantic information.} 
      ScanQA* denotes the sampled subset with scene description of ScanQA, and 3D-LLM Pre denotes the scene-alignment dataset~\cite{hong20233d}.
      }
      \label{tab:LL3DA_semantic_ablation}
    \vspace{-3mm}
\end{table*}

Based on the observations in \cref{sec:motivaiton_sub1} and \cref{sec:motivaiton_sub2}, a straightforward hypothesis is that current 3D scene encoders, often adapted from existing 3D object detection backbones for feature extraction, lack sufficient semantic information compared to 3D object-centric and 2D image-based approaches. Consequently, the pre-training stage alone is insufficient for the LLM to effectively map the extracted 3D features to the text latent space. This limitation potentially leads the model to prioritize learning patterns between questions and answers, rather than achieving genuine visual understanding, thus underutilizing the 3D tokens.

To validate the hypothesis that the 3D encoder lacks sufficient semantic information, we utilized scene descriptions from the ScanNet subset of 3D-LLM~\cite{hong20233d}. We encoded these descriptions into text embeddings using CLIP to serve as a multi-modal input representing the scene. Given that these descriptions are only available for the ScanNet training split, we sample the final 100 scenes of the train split as a test set and reconstruct the training data for ScanQA, ScanRefer, and Nr3D accordingly. As shown in \cref{tab:LL3DA_semantic_ablation}, the pre-training stage proves effective when using text embeddings as the 3D scene representation. However, pre-training with the 3D scene encoder tends to be ineffective and can even lead to performance degradation. More details please refer to \textbf{\textcolor{blue}{Supplementary Material \blueref{sec:appendix_semantic_ablation} and \blueref{sec:appendix_token_distribution}}}.

Finally, under the same experimental settings, we observe a comparable performance between using text embeddings from scene description and employing 3D tokens extracted by the 3D scene encoder. Despite the potential lack of fine-grained details in the scene descriptions, they still provide information about the object categories and their spatial relationships within the scene. Therefore, in contrast with the initial assumption, we infer that the lack of semantic information in the 3D scene encoder may not be the primary factor contributing to our earlier observations.

\subsection{Ablation of question template}

\begin{table*}[h]
	\centering
  \addtolength{\tabcolsep}{5pt}
  \setlength{\tabcolsep}{3pt}
 \begin{tabular}{ c | c | c c | c }
\toprule
\multirow{2}{*}{LLM} & \multirow{2}{*}{3D input} & \multicolumn{2}{c|}{Dataset} & \multirow{2}{*}{Accuracy$\uparrow$} \\
& & Pre-train & SFT & \\

 \midrule 
    \multicolumn{3}{l}{\textit{\textbf{Model scaling}}} \\
    \midrule

Opt-125m  & \xmark & 3D-LLM Pre & ScanQA & 35.56  \\
Qwen2-0.5B  & \xmark & 3D-LLM Pre & ScanQA & 68.15 \\
Qwen2-1.5B  & \xmark & 3D-LLM Pre & ScanQA & 88.19  \\

\midrule  
    \multicolumn{3}{l}{\textit{\textbf{Training stage}}} \\
    \midrule

\multirow{3}{*}{Qwen2-1.5B}  & \xmark & ScanQA & --- & 18.97 \\
  & \xmark & --- & ScanQA & 85.26 \\
 & \xmark & 3D-LLM Pre & ScanQA & 88.19 \\
\midrule
Qwen2-1.5B  & \cmark & 3D-LLM Pre & ScanQA & 90.65 \\

\midrule 
    \multicolumn{3}{l}{\textit{\textbf{Data scaling}}} \\
    \midrule

\multirow{4}{*}{Qwen2-1.5B}  & \xmark & 3D-LLM Pre & $\frac{1}{2}$ScanQA & 86.80 \\
& \xmark & 3D-LLM Pre & ScanQA & 88.19 \\
& \xmark & 3D-LLM Pre & ScanQA,3D-LLM QA & 91.74 \\
& \xmark & 3D-LLM Pre,ScanQA,3D-LLM QA & ScanQA,3D-LLM QA & 91.70 \\

\bottomrule

\end{tabular}
      \caption{\textbf{Analysis of experiments on ScanQA-Choice.} 3D-LLM Pre and 3D-LLM QA denotes the scene-alignment and question answering from ~\cite{hong20233d}. We leverage two-layer MLPs as a projector to avoid Q-Former directly learning the text embedding of the question.}
      \label{tab:scanqa_choice_ablation}
    \vspace{-3mm}
\end{table*}

\begin{table*}[h]
	\centering
  \addtolength{\tabcolsep}{5pt}
  \setlength{\tabcolsep}{3pt}
 \begin{tabular}{ c c | l }
\toprule
 Final loss$\downarrow$ & Accuracy(EM@1)$\uparrow$ & Ablation step \\

 \midrule  

0.2 & 89.79 & Base version of choice ScanQA \\
0.2 & 89.75 & Delete instructions \\
0.4 & 76.62 & Delete instructions and option C\\
0.4 & 75.72 & Delete instructions and option B,C\\
\rowcolor{mygray}  1.5 & 18.56 & Delete instructions and option A,B,C = Basic ScanQA setting\\

\midrule

0.2 & 89.6 & No 3D input \\

\bottomrule

\end{tabular}
      \caption{\textbf{Further analysis of instructions with MLP projector on ScanQA-Choice.} 3D VLMs with Qwen2-1.B and two MLP layers are trained under pre-train and SFT stages for 1 epoch, respectively. The gray line is equivalent to the original ScanQA, where the Accuracy metric is converted to EM@1. Assuming the correct answer is D among four options of [A,B,C,D].}
      \label{tab:scanqa_choice_further_ablation_mlp}
    \vspace{-3mm}
\end{table*}

While current 2D VLMs often evaluate performance on large, well-known datasets using a multiple-choice format, 3D VLMs still primarily rely on traditional metrics. Given the inherent complexity of spatial structures, this raises the question: \emph{would the model be providing a correct understanding that isn't accurately reflected in the evaluation results?} For instance, the answer to "What is in front of you?" varies depending on a person's orientation in space. In contrast, the multiple-choice format inherently constrains the model's predictions within the distribution of the options.

To further eliminate the influence of problem format and evaluation metrics, we propose ScanQA-Choice, a multiple-choice version of ScanQA. More visualization and detail, please refer to \textbf{\textcolor{blue}{Supplementary Material \blueref{sec:appendix_visual_scanqa_choice}}}. 
We collect the answers from the ScanQA and classify them according to the categories of answers and questions(e.g., quantity, color, object category), assigning the most similar options to each question. As shown in \cref{tab:scanqa_choice_ablation}, our experiments across various settings reveal clear benefits from model scaling, data scaling, and the pre-train and SFT stages on ScanQA-Choice, which is not evident on the original ScanQA. More details please refer to \textbf{\textcolor{blue}{Supplementary Material \blueref{sec:appendix_scanqa_choice}}}. 
However, we observed that even without providing 3D token input, the model still achieves high accuracy. This aligns with our findings in \cref{sec:motivaiton_sub1}, suggesting that the model might be leveraging memorized patterns between questions and answers learned during the SFT stage to attain higher performance. While we attempted to mitigate this by introducing question-irrelevant options in ScanQA-Choice, the overall results remained largely consistent.

We further investigated this potential "shortcut" through ablation studies, as shown in \cref{tab:scanqa_choice_further_ablation_mlp}. We observed that instruction prompts are not critical, suggesting the model could inherently learn the relationship between questions and answers. In contrast, the number of answer options proved to be a significant factor. Progressively reducing the number of options led to lower convergence and poorer final performance, with a drastic drop occurring when the last choice was removed. Additional experiments exploring the impact of providing supplementary information in \textbf{\textcolor{blue}{Supplementary Material \blueref{sec:appendix_scanqa_choice}}}, such as answer length or random options, indicated a marginal but non-essential contribution to the performance.

In summary, our findings confirm that under the current data scale, employing a multiple-choice QA format is not optimal. The model tends to disregard the 3D tokens and instead focuses on learning the newly provided information within the options.

\subsection{Ablation of data distribution}
\label{sec:ablation_data_distribution}

\begin{table*}[h]
	\centering
  \addtolength{\tabcolsep}{5pt}
  \setlength{\tabcolsep}{3pt}
 \begin{tabular}{ c c | c | c c c }
\toprule
 Train set & Test set & Data balance  & BLUE-4 $\uparrow$ & CIDEr$\uparrow$ & ROUGE $\uparrow$ \\

 \midrule  
    \multicolumn{3}{l}{\textit{\textbf{Dataset distribution}}} \\
    \midrule

 \colorbox{mycyan}\cmark & \xmarkg & \xmarkg & 11.83 & 75.39 & 38.12 \\
  \cmarkg & \colorbox{mycyan}\cmark & \xmarkg & 13.26 & 79.06 & 40.36 \\
  \colorbox{mypink}\xmarkg & \cmarkg & \xmarkg & 14.09 & 82.66 & 40.88 \\

\midrule  
    \multicolumn{3}{l}{\textit{\textbf{Distribution balancing}}} \\
    \midrule

 \colorbox{mycyan}\cmark & \xmarkg & \xmarkg & 11.83 & 75.39 & 38.12 \\
 \cmarkg & \xmarkg & \colorbox{mycyan}\cmark & 9.42 & 64.47 & 33.63 \\

\bottomrule

\end{tabular}
      \caption{\textbf{Analysis of GT data distribution.} Leveraging Qwen2-1.5B as LLM backbone, experiments are conducted on ScanQA with pre-training and SFT stages.}
      \label{tab:LL3DA_test_set_ablation}
    \vspace{-6mm}
\end{table*}

\begin{figure*}[h]
    \centerline{\includegraphics[width=\textwidth]{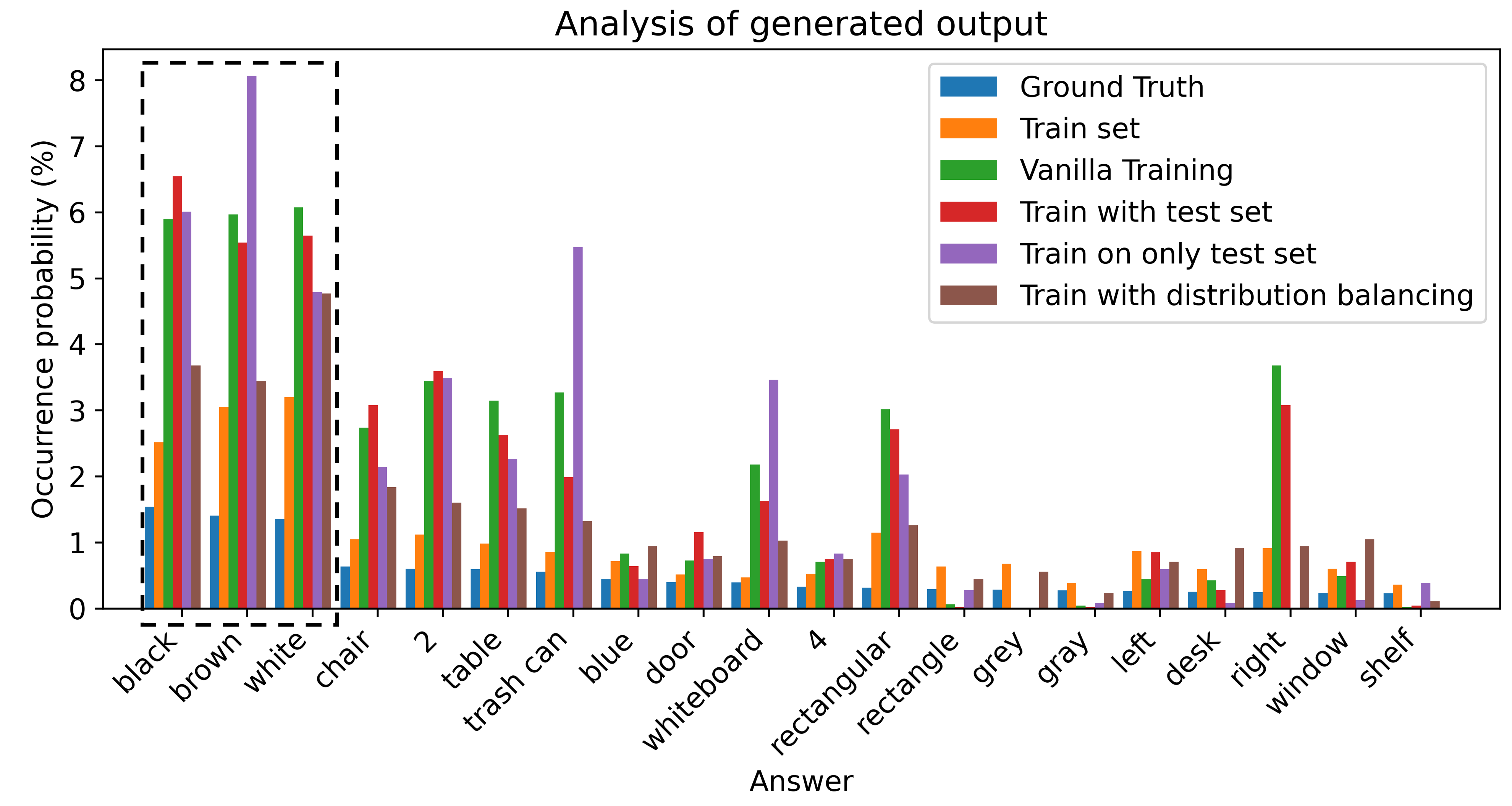}}
    \caption{\textbf{Analysis of generated answer frequency.} The top 20 generated answers under various training settings show that test data inclusion did not improve fitting to frequent answers and might generalize them to other questions. }
    \vspace{-3mm}
    \label{fig:analysis_generated_answer_frequency}
\end{figure*}

Having excluded factors of semantic information and question format, we observed that the model excessively relies on textual questions to model the relationship between questions and answers. This motivated us to further investigate the distribution of GT answers in the dataset. As shown in \cref{tab:LL3DA_test_set_ablation}, we conducted experiments involving the training dataset and data balancing. The results indicate that incorporating the test set into the current training setting yields a slight improvement, which is in line with our expectations, while applying data balancing led to a degradation in model performance. 
Consequently, as illustrated in \cref{fig:analysis_generated_answer_frequency}, we further visualized the answer occurrence probabilities. We sort the answers based on their frequency in the test set and visualize the predicted answer distributions under various settings. More details of the Top50 generated output, please refer to \textbf{\textcolor{blue}{Supplementary Material \blueref{sec:further_generated_output}}}. 
It is evident that on ScanQA, the predicted answers are significantly concentrated on the most frequent answers (within the black dashed box), with a probability much higher than their occurrence in the test set. Simultaneously, we observed that data balancing effectively suppresses the overfitting to these high-frequency answers. However, as depicted in \cref{fig:analysis_generated_answer_frequency}, while this benefits the model's generalization ability, it does not necessarily enhance its overall performance.

\subsection{Summary and Verification}

\begin{table*}[h]
	\centering
  \addtolength{\tabcolsep}{5pt}
  \setlength{\tabcolsep}{3pt}
 \begin{tabular}{ c | c | c c | c }
\toprule
LLM & 3D input & Pre-train & SFT  & Accuracy $\uparrow$ \\

\midrule 

\multirow{3}{*}{Qwen2-1.5B} & \xmarkg & \xmarkg & \cmarkg & 0 \\
 & \colorbox{mycyan}\cmark & \xmarkg & \cmarkg & 58.95 \\
 & \cmarkg & \colorbox{mycyan}\cmark & \cmarkg & 76.26 \\

\bottomrule

\end{tabular}
      \caption{\textbf{Verification on our designed 3D-RDQA dataset.} Two MLP layers are adopt for projector.}
      \label{tab:final_verification}
    \vspace{-3mm}
\end{table*}

\textbf{Summary.} In brief, we can answer the questions raised in \cref{sec:motivation}: 3D VLMs are not inherently incapable of utilizing 3D tokens. However, the imbalance within the datasets leads the model to heavily rely on input text to generate fixed responses for better performance, thereby obviating the need to consider 3D spatial information. This phenomenon suggests that the model does not genuinely see the 3D space. Consequently, the 3D encoder weights or even the 3D Encoder itself become dispensable. Similarly, the pre-training alignment intended to facilitate better utilization of the 3D Encoder becomes unnecessary. Finally, the inconsistency in QA distributions across different datasets explains why performance gains observed on small-scale datasets do not scale up to larger datasets. This also highlights that the observed issue is likely not isolated to 3D scene-centric VLMs, but rather a potential challenge inherent to all 3D VLMs.

\textbf{Verification.} To further validate our finding with considering that modifying the model architecture can be complex and potentially harm generalization, we adopted a data-centric approach by designing a simple yet effective strategy. Our core idea is to ensure that 3D tokens influence the final answer. Specifically, based on our designed ScanQA-Choice, we "poison" the 3D tokens of each 3D-QA pair to create a new, modified QA pair. This modified pair, along with the original QA pair, forms a 3D Relevance Discrimination QA (3D-RDQA) pair. For more details about 3D-RDQA, please refer to \textbf{\textcolor{blue}{Supplementary Material \blueref{sec:appendix_3DRDQA}}}. This strategy offers two main benefits: first, if the model lacks 3D perception, it will be unable to distinguish between the original and modified QA, thus disrupting its reliance on question cues and revealing its true capabilities. Second, this encourages the model to recognize and understand the differences conveyed by 3D tokens, thereby improving its 3D spatial understanding. 
As shown in \cref{tab:final_verification}, we successfully observed the impact of the 3D encoder and the improvements brought by the pre-training stage based on our designed 3D-RDQA dataset, which validates the correctness of our findings.

%% file: sec/5_conclusion.tex
\section{Conclusion}
In this work, we identify three key differences between 3D scene-centric and 2D VLMs concerning semantic understanding, question format requirements, and data distribution. We find that current 3D datasets suffer from repetitive patterns, causing models to overfit text rather than learn true 3D spatial reasoning. To address this, we introduce the 3D-RDQA dataset, designed to break such shortcuts and encourage spatial understanding. This dataset facilitates more rigorous evaluation and future progress towards enhanced spatial reasoning in 3D VLMs.

\textbf{Limitations.} While the 3D-RDQA dataset effectively evaluates the 3D understanding of models, its multiple-choice format requires further investigation of its adaptability to other tasks. The core principle of 3D-RDQA lies in contrasting model behavior across diverse data. Therefore, we believe that approaches like Direct Preference Optimization (DPO) hold promise for demonstrating even stronger performance and providing a more direct evaluation of 3D understanding.

%% file: sec/X_suppl.tex
\appendix
\newpage

\section{Appendix/supplemental material}
The outline of the Appendix is as follows:
\begin{itemize}
    \item More implementation details;
    \item More analysis on semantic information;
    \item More analysis on the ScanQA-Choice;
    \item More analysis on data scaling capabilities on 3D-DC;
    \item More visualization of token distribution;
    \item More example visualization of ScanQA-Choice;
    \item More visualization of generated output;
    \item More example visualization of 3D-RDQA dataset;
    \item More discussion;
        \begin{itemize}
            \item Discussion on performance analysis of 3D scene-centric VLM;
            \item Discussion on performance analysis on 3D-RDQA dataset;
            \item Discussion on performance analysis compared with 3D object-centric methods;
            \item Discussion on performance analysis of 3D-VG;
            \item Discussion on expansion to 3D-DC and 3D-VG with relevance discrimination idea;
            \item Discussion on social impact;
        \end{itemize}
\end{itemize}

\section{Implementation Details}
\textbf{Datasets.} Following ~\cite{chen2024ll3da}, we utilize the ScanNet dataset~\cite{dai2017scannet}, which comprises 1,201 and 312 diverse and complex indoor 3D scenes for training and validation, respectively. By default, experiments are conducted with the same setting with ~\cite{chen2024ll3da} no ScanQA~\cite{azuma2022scanqa}, ScanRefer~\cite{chen2020scanrefer}, Nr3D~\cite{achlioptas2020referit3d} and the ScanNet subset of 3D-LLM~\cite{hong20233d}. We further divide the ScanNet subset of 3D-LLM into two parts: 3D-LLM QA and 3D-LLM Pre. The 3D-LLM Pre subset encompasses scene descriptions, conversations, and embodied planning tasks.

\textbf{Metrics.} We adopt C, B-4, R as abbreviations for CIDEr~\cite{vedantamconsensus}, BLEU-4~\cite{papineni2002bleu}, and Rouge-L~\cite{lin2004rouge} to evaluate the quality of the generated textual responses, while accuracy and EM@1 is leveraged to evaluate quality under multiple-choice dataset.

\textbf{Implementation Details.} We follow~\cite{chen2024ll3da} to sample 40k point clouds from each scene for 3D scene encoder~\cite{chen2024vote2cap}. We leverage open-source Qwen2-1.5B~\cite{yang2024qwen2} as LLM backbone and Q-Former~\cite{li2023blip} as projector by default. Following ~\cite{liu2023visual}, we train model by pre-train and SFT stages for 1 epoch, respectively. We adopt AdamW~\cite{loshchilov2017decoupled} as optimizer with a weight decay of 0.1 and a learning rate decaying from 10$^{-4}$ to 10$^{-5}$ with a cosine annealing scheduler for pre-train stage, while a learning rate decaying from 5$\times$10$^{-5}$ to 10$^{-5}$ is leveraged for SFT stage. For all the training tasks, we train with a total batch size of 32 on 8$\times$Ascend-D910 (64G) NPU. We observe that training for only one epoch in both the pre-train and SFT stages without gradient clipping, can yield comparable performance to that achieved by LL3DA training Q-Former for 32 epochs.

\section{Additional analysis of semantic information}
\label{sec:appendix_semantic_ablation}

\begin{table*}[h]
	\centering
  \addtolength{\tabcolsep}{5pt}
  \setlength{\tabcolsep}{3pt}
 \begin{tabular}{ c c c | c c c }
\toprule
Multi-modal input & Pre-train & 3D object token  & BLUE-4 $\uparrow$ & CIDEr$\uparrow$ & ROUGE $\uparrow$ \\
\midrule 

 \multirow{4}{*}{\text{Scene description}} & ScanQA* & \xmark & 5.18 & 72.42 & 26.68 \\
 & 3D-LLM Pre & \xmark & 5.40 & 75.74 & 27.78 \\

 & ScanQA* & \cmark & 4.85 & 72.67 & 26.84 \\
 & 3D-LLM Pre & \cmark & 5.25 & 75.47 & 27.79 \\

\midrule

\multirow{4}{*}{\text{3D scene}} & ScanQA* & \xmark & 5.37 & 77.55 & 28.35 \\
 & 3D-LLM Pre & \xmark & 5.54 & 76.30 & 27.92 \\

 & ScanQA* & \cmark & 5.33 & 77.59 & 28.44 \\
 & 3D-LLM Pre & \cmark & 4.90 & 74.40 & 27.48 \\

\bottomrule

\end{tabular}
      \caption{\textbf{Analysis of semantic information.} 
      ScanQA* denotes the sampled subset with scene description of ScanQA and 3D-LLM Pre denotes the scene-alignment dataset~\cite{hong20233d}.
      }
      \label{tab:appendix_LL3DA_semantic_ablation}
\end{table*}

As shown in \cref{tab:appendix_LL3DA_semantic_ablation}, our analysis on training augmentations, specifically the inclusion of 3D tokens inside the GT bounding boxes for potential object-level 3D representations, indicate that this augmentation has a minimal impact on the model's performance.

\section{Additional analysis of ScanQA-Choice}
\label{sec:appendix_scanqa_choice}

\begin{table*}[h]
	\centering
  \addtolength{\tabcolsep}{5pt}
  \setlength{\tabcolsep}{3pt}
 \begin{tabular}{ c | c | c c | c }
\toprule
\multirow{2}{*}{LLM} & \multirow{2}{*}{3D input} & \multicolumn{2}{c|}{Dataset} & \multirow{2}{*}{Accuracy} \\
& & Pre-train & SFT & \\

\midrule 
    \multicolumn{3}{l}{\textit{\textbf{Data scaling}}} \\
    \midrule

\multirow{4}{*}{Qwen2-0.5B} & \multirow{4}{*}{\xmark} & 3D-LLM Pre & $\frac{1}{2}$ScanQA & 77.65\textcolor{orange}{(-0.9)}  \\
 &  & 3D-LLM Pre & ScanQA & 78.55 \\
 &  & 3D-LLM Pre & ScanQA,3D-LLM QA & 90.48 \textcolor{blue}{(+11.93)} \\
 &  & 3D-LLM Pre\&QA,ScanQA & ScanQA,3D-LLM QA & 90.93 \textcolor{blue}{(+0.45)}\\

\midrule

 \multirow{4}{*}{Qwen2-1.5B} & \multirow{4}{*}{\xmark} & 3D-LLM Pre & $\frac{1}{2}$ScanQA & 89.26 \textcolor{orange}{(-1.39)} \\
 &  & 3D-LLM Pre & ScanQA & 90.65 \\
 &  & 3D-LLM Pre & ScanQA,3D-LLM QA & 91.74 \textcolor{blue}{(+1.09)} \\
 &  & 3D-LLM Pre\&QA,ScanQA & ScanQA,3D-LLM QA & 91.70 \textcolor{orange}{(-0.04)}\\

\bottomrule

\end{tabular}
      \caption{\textbf{Further analysis on ScanQA-Choice.} We supplement experiments with Qwen2-0.5B, which can better perform data scaling capabilities. 3D-LLM Pre denotes the scene-alignment dataset~\cite{hong20233d}. We leverage two layer MLPs as projector to avoid Q-Former directly learning text embedding of question. (*) denotes performance change compared to the basic setting of leveraging 3D-LLM Pre for pre-training and ScanQA for SFT.}
      \label{tab:scanqa_choice_further_ablation_model}
\end{table*}

We supplement experiments on ScanQA-Choice with different LLM backbone. As shown in \cref{tab:scanqa_choice_further_ablation_model}, the benefits of data scaling are more pronounced when using smaller models, such as Qwen2-0.5B. This suggests that with a larger model like Qwen2-1.5B, performance may have approached saturation on smaller datasets without 3D input.

\begin{table*}[h]
	\centering
  \addtolength{\tabcolsep}{5pt}
  \setlength{\tabcolsep}{3pt}
 \begin{tabular}{  c c | l }
\toprule
Final loss$\downarrow$ & Accuracy(EM@1)$\uparrow$ & Ablation step \\

 \midrule 

0.3 & 90.82 & Base version of choice ScanQA \\
0.5 & 84.45 & Delete instructions \\
0.6 & 75.61 & Delete instructions and option C\\
0.8 & 65.25 & Delete instructions and option B,C\\
\rowcolor{mygray} 1.7 & 15.67 & Delete instructions and option A,B,C = Basic ScanQA setting\\

\midrule

1.5 & 23.49 & Only provide the length of the answer\\
1.7 & 17.98 & Randomly sample a question as an option\\

\bottomrule

\end{tabular}
      \caption{\textbf{Further analysis of instructions with Q-Former.} 3D VLMs with Qwen2-1.5B and Q-Former are only trained under pre-train stage for 1 epoch. The gray line is equivalent to the original ScanQA, where the Accuracy metric is converted to EM@1. Assuming the correct answer is D among four options of [A,B,C,D].}
      \label{tab:appendix_scanqa_choice_further_ablation_qformer}
\end{table*}

As shown in \cref{tab:appendix_scanqa_choice_further_ablation_qformer}, we supplement further analysis on ScanQA-Choice with Q-Former. When employing Q-Former as the projector, the model, likely due to Q-Former's text processing capabilities, achieves high performance even without the SFT stage. 

\section{Additional analysis of data scaling capabilities on 3D-DC}
\label{sec:appendix_data_scaling_densecap}

\begin{table*}[h]
	\centering
  \setlength{\tabcolsep}{3pt}
 \begin{tabular}{ c | c c c c | c c c }
\toprule
\multirow{2}{*}{LLM} & \multicolumn{2}{c}{3D-DC} & 3D-QA & Pre-train & \multirow{2}{*}{BLUE-4 $\uparrow$} & \multirow{2}{*}{CIDEr$\uparrow$} & \multirow{2}{*}{ROUGE $\uparrow$} \\

  & Nr3D & ScanRefer & ScanQA & 3D-LLM & & & \\
 
 \midrule 
    \multicolumn{3}{l}{\textit{\textbf{Official LL3DA}}} \\
    \midrule
  Opt-1.3B & \cmark & \cmark & \cmark & \cmark & \textcolor{lightgray}{13.37} & \textcolor{lightgray}{23.94} & \textcolor{lightgray}{45.78} \\

 \midrule 
    \multicolumn{3}{l}{\textit{\textbf{Scaling with other task}}} \\
    \midrule

\multirow{3}{*}{Qwen2-1.5B} & \colorbox{mycyan}{\cmark} & \xmarkg & \xmarkg & \xmarkg & 23.12 & 38.58 & 52.00 \\
 & \cmarkg & \xmarkg & \colorbox{mycyan}{\cmark} & \xmarkg & 22.89\textcolor{orange}{(-0.23)} & 36.86\textcolor{orange}{(-1.72)} & 51.29\textcolor{orange}{(-0.71)} \\
 & \cmarkg & \xmarkg & \cmarkg & \colorbox{mycyan}{\cmark} & 23.62\textcolor{blue}{(-0.40)} & 39.12\textcolor{blue}{(+0.54)} & 51.69\textcolor{orange}{(-0.31)} \\

\midrule 
    \multicolumn{3}{l}{\textit{\textbf{Scaling with 3D-DC}}} \\
    \midrule

\multirow{5}{*}{Qwen2-1.5B} & \colorbox{mycyan}{\cmark} & \xmarkg & \xmarkg & \xmarkg & 23.12 & 38.58 & 52.00 \\
 & \cmarkg & \colorbox{mycyan}{\cmark} & \xmarkg & \xmarkg & 12.05\textcolor{orange}{(-11.07)} & 18.23\textcolor{orange}{(-20.35)} & 42.94\textcolor{orange}{(-9.06)} \\
& \cmarkg & \cmarkg & \colorbox{mycyan}{\cmark} & \xmarkg & 11.55\textcolor{orange}{(-11.57)} & 18.21\textcolor{orange}{(-20.37)} & 42.88\textcolor{orange}{(-9.12)} \\
 & \cmarkg & \cmarkg & \cmarkg & \colorbox{mycyan}{\cmark} & 11.64\textcolor{orange}{(-11.48)} & 18.29\textcolor{orange}{(-20.29)} & 42.54\textcolor{orange}{(-9.46)} \\

\bottomrule


\end{tabular}
      \caption{\textbf{Further analysis of data scaling capabilities on Nr3D.} Following our investigation into the data scaling capabilities for the 3D Dense Captioning task with Nr3D-centric setting, our further analysis reveals that Nr3D does not benefit from data scaling. On the contrary, it potentially leads to a degradation in performance. 3D-LLM denotes the scene-alignment dataset~\cite{hong20233d}. (*) denotes performance change compared to train only on Nr3D.}
      \label{tab:further_data_scaling_ablation_densecap_nr3d}
\end{table*}

\begin{table*}[h]
	\centering
  \setlength{\tabcolsep}{3pt}
 \begin{tabular}{ c | c c c c | c c c }
\toprule
\multirow{2}{*}{LLM} & \multicolumn{2}{c}{3D-DC} & 3D-QA & Pre-train & \multirow{2}{*}{BLUE-4 $\uparrow$} & \multirow{2}{*}{CIDEr$\uparrow$} & \multirow{2}{*}{ROUGE $\uparrow$} \\

  & ScanRefer & Nr3D & ScanQA & 3D-LLM   & & & \\
 
 \midrule 
    \multicolumn{3}{l}{\textit{\textbf{Official LL3DA}}} \\
    \midrule
  Opt-1.3B & \cmark & \cmark & \cmark & \cmark & \textcolor{lightgray}{35.97} & \textcolor{lightgray}{62.98} & \textcolor{lightgray}{54.65} \\

 \midrule 
    \multicolumn{3}{l}{\textit{\textbf{Scaling with other task}}} \\
    \midrule

\multirow{3}{*}{Qwen2-1.5B} & \colorbox{mycyan}{\cmark} & \xmarkg & \xmarkg & \xmarkg & 32.42 & 53.57 & 50.84 \\
 & \cmarkg & \xmarkg & \colorbox{mycyan}{\cmark} & \xmarkg & 33.60\textcolor{blue}{(+1.18)} & 54.51\textcolor{blue}{(+0.94)} & 51.33\textcolor{blue}{(+0.49)} \\
 & \cmarkg & \xmarkg & \cmarkg & \colorbox{mycyan}{\cmark} & 33.24\textcolor{blue}{(+0.82)} & 56.51\textcolor{blue}{(+2.94)} & 51.00\textcolor{blue}{(+0.16)} \\

\midrule 
    \multicolumn{3}{l}{\textit{\textbf{Scaling with 3D-DC}}} \\
    \midrule

\multirow{5}{*}{Qwen2-1.5B} & \colorbox{mycyan}{\cmark} & \xmarkg & \xmarkg & \xmarkg & 32.42 & 53.57 & 50.84 \\
& \cmarkg & \colorbox{mycyan}{\cmark} & \xmarkg & \xmarkg & 33.62\textcolor{blue}{(+1.12)} & 54.51\textcolor{blue}{(+0.94)} & 51.55\textcolor{blue}{(+0.71)} \\
& \cmarkg & \cmarkg & \colorbox{mycyan}{\cmark} & \xmarkg & 33.00\textcolor{blue}{(+0.58)} & 55.13\textcolor{blue}{(+1.56)} & 51.22\textcolor{blue}{(+0.38)} \\
& \cmarkg & \cmarkg & \cmarkg & \colorbox{mycyan}{\cmark} & 33.26\textcolor{blue}{(+0.84)} & 56.16\textcolor{blue}{(+2.58)} & 51.31\textcolor{blue}{(+0.47)} \\

\bottomrule


\end{tabular}
      \caption{\textbf{Further analysis of data scaling capabilities on ScanRefer.} Following our investigation into the data scaling capabilities for the 3D-DC task with ScanRefer-centric setting, our further analysis reveals that ScanRefer benefits from data scaling with pre-train dataset, 3D-QA and 3D-DC datasets. 3D-LLM Pre denotes the scene-alignment dataset~\cite{hong20233d}. (*) denotes performance change compared to train only on ScanRefer.}
      \label{tab:further_data_scaling_ablation_densecap_scanrefer}
\end{table*}

\begin{table*}[h]
	\centering
 \begin{tabular}{ c | c c |c c c | c c c }
\toprule
\multirow{2}{*}{LLM} & \multicolumn{2}{c|}{Role isolation} & \multicolumn{3}{c|}{Nr3D} & \multicolumn{3}{c}{ScanRefer} \\
 & QA template & Prompt prefix & B-4 $\uparrow$ & C$\uparrow$ & R $\uparrow$ & B-4 $\uparrow$ & C$\uparrow$ & R $\uparrow$ \\
 
 \midrule 

\multirow{4}{*}{Qwen2-1.5B} & \xmarkg & \xmarkg & 12.05 & 18.23 & 42.94 & 33.62 & 54.51 & 51.55 \\
 & \colorbox{mycyan}{\cmark} & \xmarkg & 17.79 & 31.92 & 46.92 & 30.59 & 51.89 & 49.20 \\
 & \cmarkg & \colorbox{mycyan}{\cmark} & 17.54 & 28.79 & 46.70 & 31.23 & 52.62 & 49.44 \\

\bottomrule


\end{tabular}
      \caption{\textbf{Analysis of role isolation for 3D VLM.} Further investigation involved the integration of isolation mechanisms to address potential conflicts observed in the Nr3D and ScanRefer datasets. 
      While the implementation of this technique facilitated a more balanced performance profile across the two datasets, it did not ultimately yield peak performance in either individual evaluation.
      }
      \label{tab:further_data_scaling_ablation_densecap_both}
\end{table*}

As shown in \cref{tab:further_data_scaling_ablation_densecap_nr3d} and \cref{tab:further_data_scaling_ablation_densecap_scanrefer}, we further supplement analysis of data scaling capabilities on 3D Dense Captioning task. As shown in \cref{tab:further_data_scaling_ablation_densecap_nr3d}, we observed that Nr3D does not benefit from data scaling from other tasks, nor does it experience a degradation in its original performance. However, we found a catastrophic performance decline on Nr3D when incorporating the ScanRefer dataset, which also focuses on 3D-DC. 
Analysis of the model's generated outputs during evaluation reveals that ScanRefer contains a high frequency of similar location descriptions starting with "it is to the". This prevalent phrase leads the model to generate such descriptions even on the Nr3D dataset, consequently impacting performance. This observation aligns with our findings regarding data distribution discussed in \cref{sec:ablation_data_distribution}. However, as shown in \cref{tab:further_data_scaling_ablation_densecap_scanrefer}, ScanRefer-centric analysis yields a contrasting conclusion: ScanRefer demonstrates effective data scaling, showing performance improvements across both similar and dissimilar tasks. 

As shown in \cref{tab:further_data_scaling_ablation_densecap_both}, to further investigate whether role isolation could mitigate the conflicts between datasets, we explored adding dataset-specific prefixes to questions and using distinct question templates for each dataset. While this approach offers some relief, we observed that the performance after role isolation consistently ends up being worse than the best performance achieved on the original datasets separately. Thus, role isolation appears to represent a trade-off rather than a definitive solution.

\section{Additional visualization of token distribution}
\label{sec:appendix_token_distribution}
\begin{figure}[h]
    \centering
    \subfigure[Visualization with scene description input]{\includegraphics[width=0.48\textwidth]{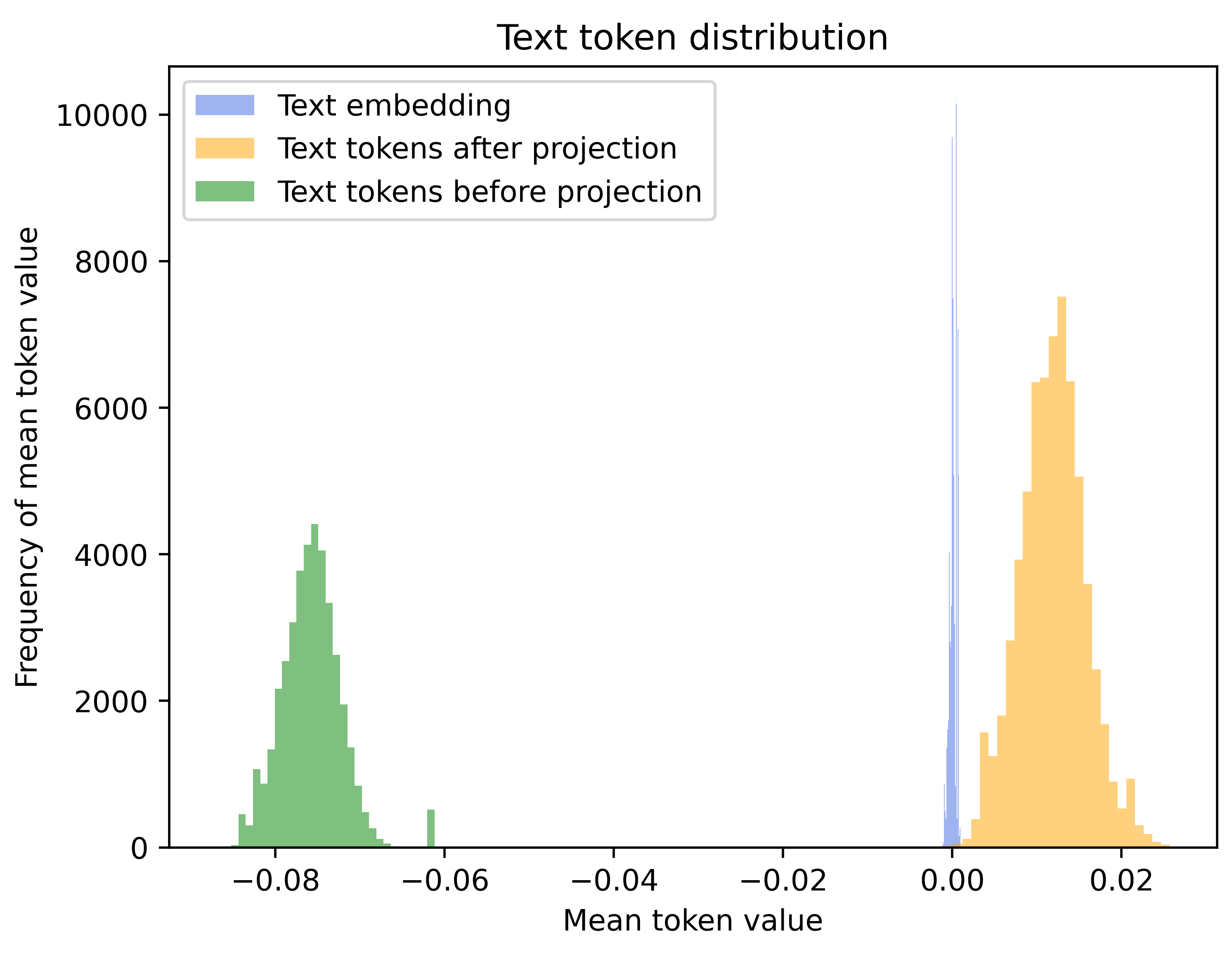}\label{fig:subfig2}}
    \subfigure[Visualization with 3D scene input]{\includegraphics[width=0.48\textwidth]{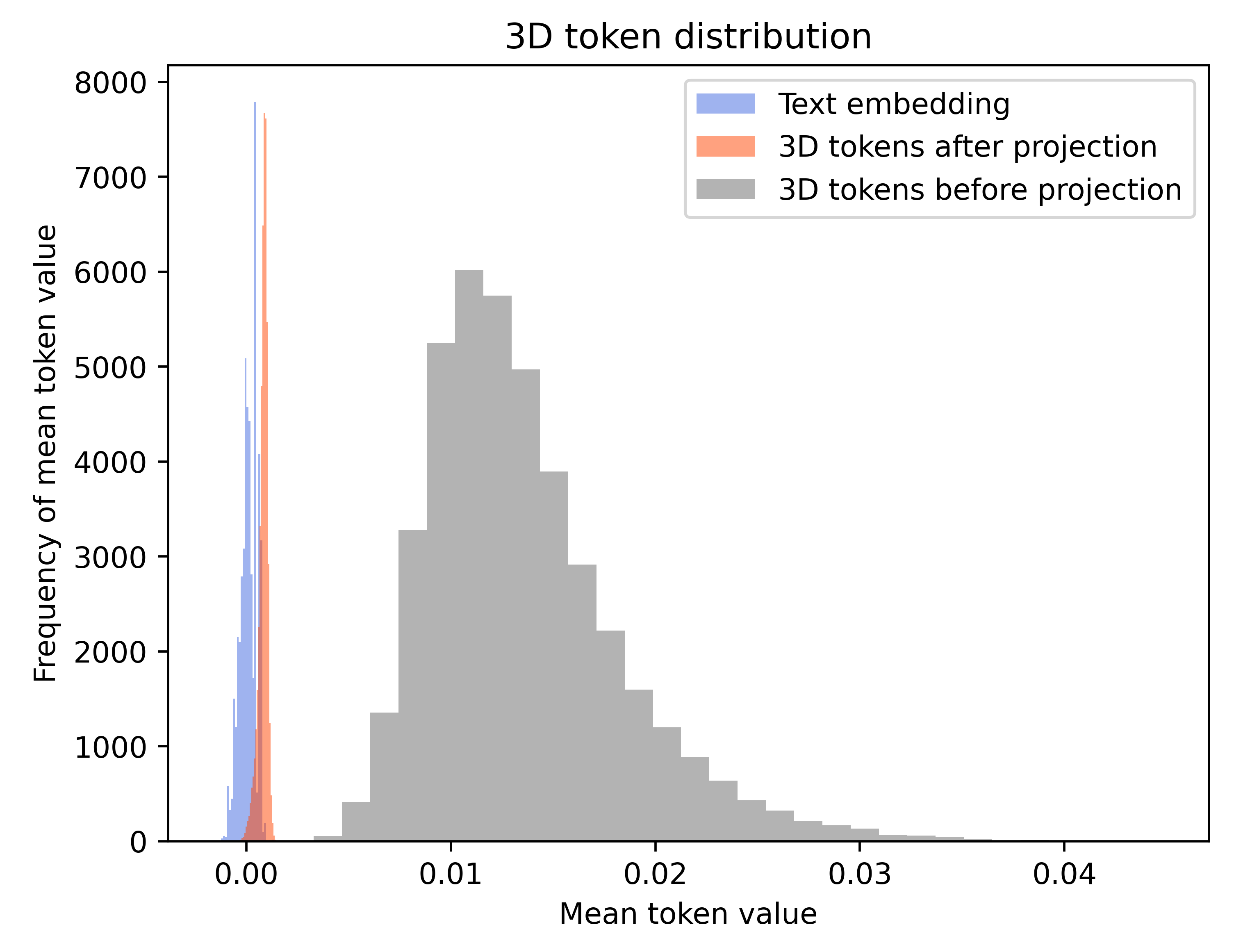}\label{fig:subfig1}}
    \caption{\textbf{Visualization of token distribution with different cross-modal input.} We further visualize the token distribution before and after MLP projector to intuitively express the impact of pre-train stage.} 
    \label{fig:token_distribution}
\end{figure}

To better understand the distribution of tokens for alignment analysis on pre-train stage, we collect tokens from the ScanQA training set both before and after the projector, compared to tokens of text embeddings. For each token, we calculate its mean vector and then visualized the distribution of these mean vectors using histograms. This allows for a comparison of how the projector influences the token representations.

As shown in \cref{fig:token_distribution}, we further visualize the distribution of text tokens from scene descriptions with CLIP and 3D tokens from the 3D encoder before and after the pre-train stage. While pre-training aims to align disparate data distributions with text for better feature learning, our visualization surprisingly shows that the 3D encoder effectively maps 3D tokens to a distribution even closer to text than using text tokens. This indicates the model's underlying capability to utilize 3D tokens.

\section{Additional example visualization of ScanQA-Choice}
\label{sec:appendix_visual_scanqa_choice}
\begin{figure*}[h]
    \centerline{\includegraphics[width=\textwidth]{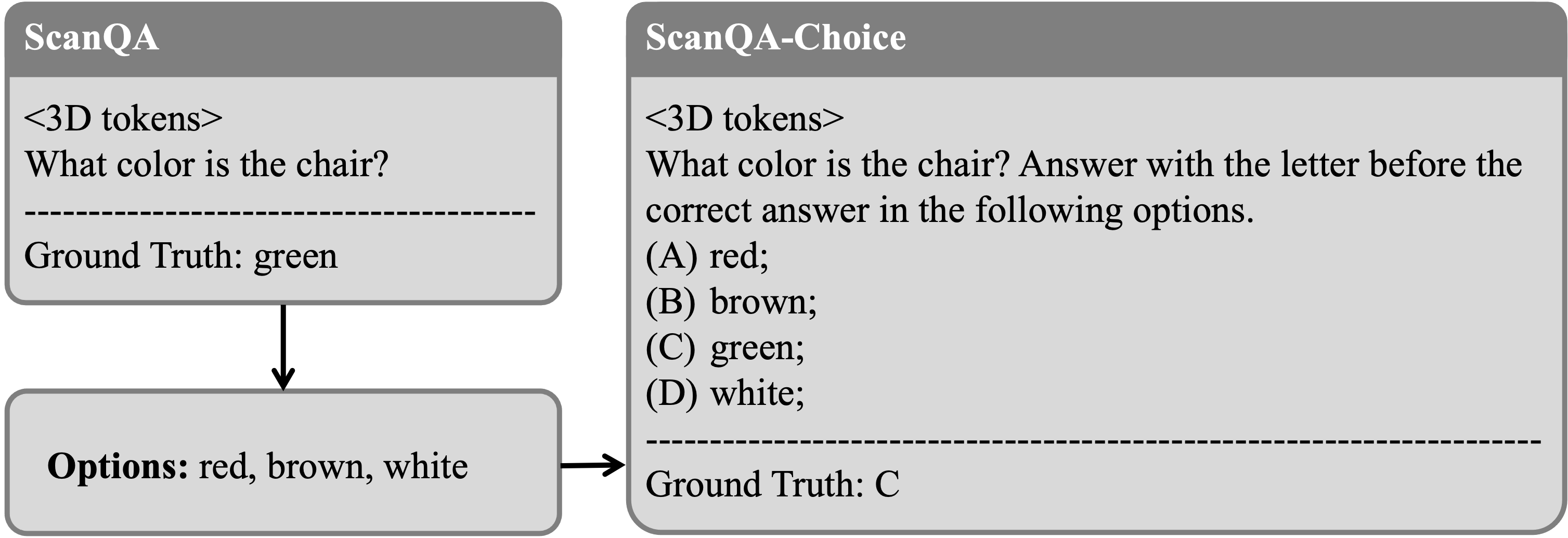}}
    \caption{\textbf{Example visualization of ScanQA and ScanQA-Choice collection.} Based on the ground truth answer for each question in ScanQA, we sampled similar options from the ScanQA answer pool to construct ScanQA-Choice.
    }
    \label{fig:visualization_scanqa_choice}
\end{figure*}

As shown in \cref{fig:visualization_scanqa_choice}, we present a visual demonstration of how ScanQA-Choice is constructed based on ScanQA.

\section{Additional visualization of generated output}
\label{sec:further_generated_output}
\begin{figure*}[h]
    \centerline{\includegraphics[width=\textwidth]{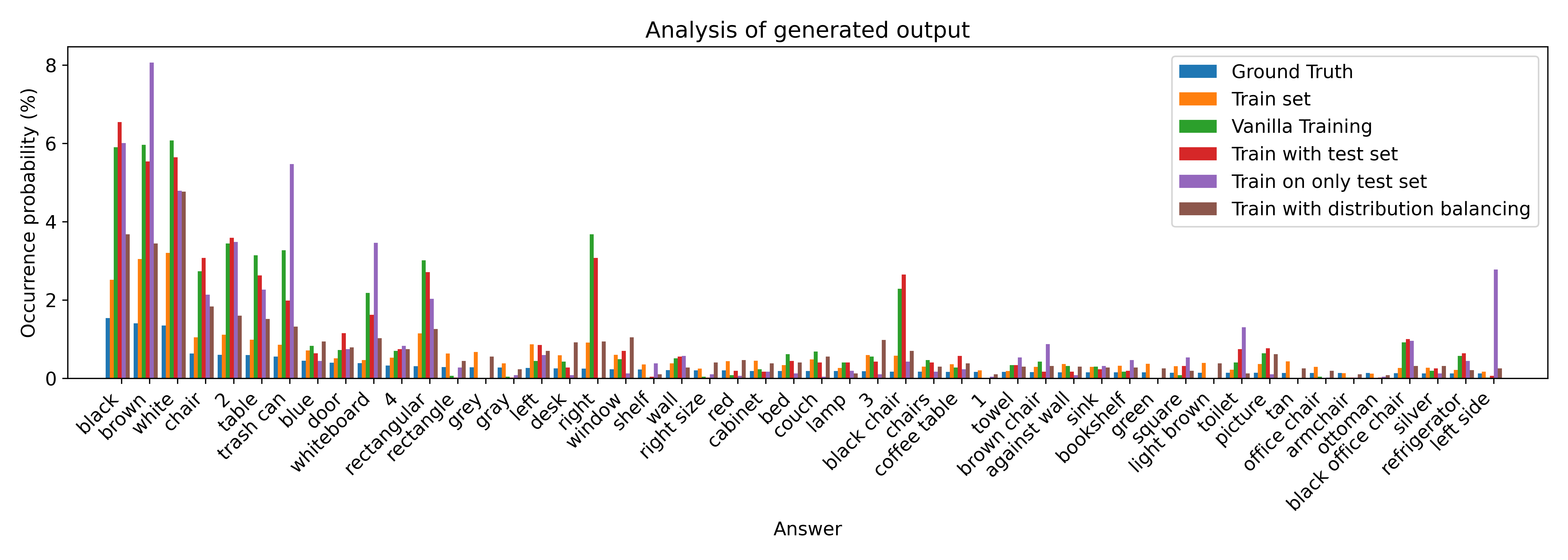}}
    \caption{\textbf{Further analysis of generated answer frequency.} The top 50 generated answers under various training settings. }
    \vspace{-3mm}
    \label{fig:further_analysis_generated_answer_frequency}
\end{figure*}

As shown in \cref{fig:further_analysis_generated_answer_frequency}, we augmented the top 50 generated answers and observed that, while including the test set in training generally improves evaluation metrics, the answer distribution does not necessarily improve and can even worsen, as seen with answers like "brown chair" and "toilet." Furthermore, the generated answers within the Top 50 occurrence probability exhibit higher frequencies than the ground truth, suggesting poorer performance on questions with less frequent answers.

\section{Additional example visualization of 3D-RDQA dataset}
\label{sec:appendix_3DRDQA}
\begin{figure}[h]
    \centerline{\includegraphics[width=\textwidth]{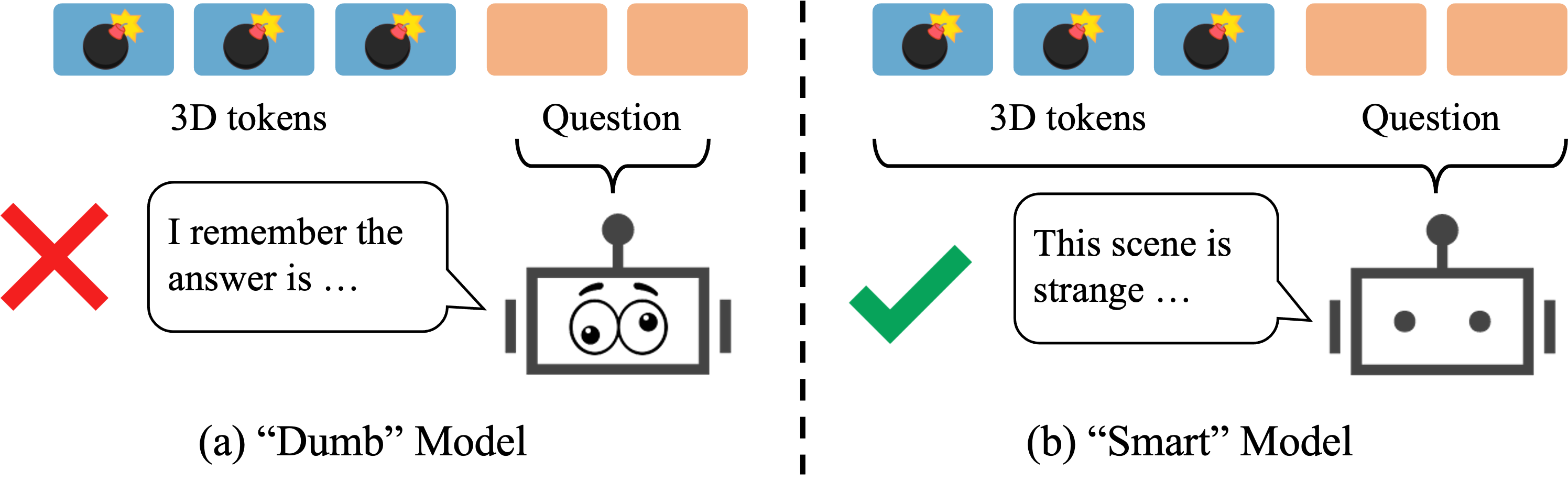}}
    \caption{\textbf{Model Comparison:} (a) A "dumb" model ignores 3D tokens, relying only on text, (b) A "smart" model understands 3D tokens and their relation to text. 3D tokens with "bomb" denotes the poisoned 3D tokens.} 
    \label{fig:relevance_discrimination_example}
\end{figure}

\begin{figure*}[h]
    \centerline{\includegraphics[width=\textwidth]{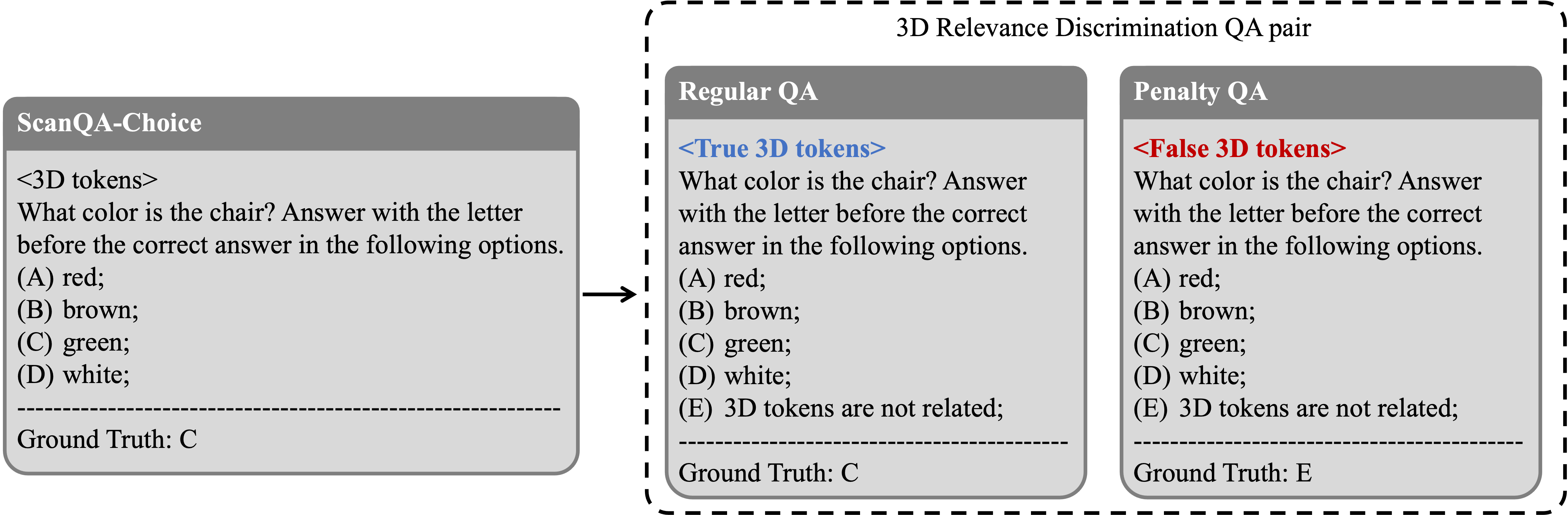}}
    \caption{\textbf{Example visualization of 3D-RDQA pair collection.} Utilizing our constructed ScanQA-Choice dataset, we generate a 3D-RDQA pair by modifying 3D tokens and introducing a novel choice option.}
    \vspace{-3mm}
    \label{fig:relevance_discrimination_qa}
\end{figure*}

As shown in \cref{fig:relevance_discrimination_example}, we first visualize the motivation for constructing the 3D-RDQA dataset. When "poisoning" 3D tokens, 3D VLMs heavily reliant on text tend to disregard the changes due to a lack of 3D scene understanding. In contrast, 3D VLMs with genuine 3D spatial reasoning can clearly identify the mismatch between the 3D tokens and the question-answer pair.

Moreover, as shown in \cref{fig:relevance_discrimination_qa}, we primarily generate 3D-RDQA pairs by manipulating the correspondence between 3D tokens and the question-answer pair. Specifically, we can efficiently obtain \textcolor{red}{<False 3D tokens>} by simply ensuring the loaded 3D tokens are sourced from a different scene.


\section{Discussion}
\subsection{Performance analysis of 3D scene-centric VLM}
Our analysis of 3D scene-centric VLMs aligns with the findings of this paper. First, 3D-LLM~\cite{hong20233d} performance is notably weak compared with other 3D scene-centric approaches, even falling below that of models without a 3D encoder in this paper, likely due to differences in training methodologies. Second, Grounded 3D-LLM~\cite{chen2024grounded}, despite significant effort in training an object-alignment scene encoder, shows limited performance gains on ScanQA, consistent with our observations in ~\cref{sec:motivaiton_sub2}. Finally, LSceneLLM~\cite{zhi2024lscenellm} achieves improved ScanQA performance through finer-grained feature selection. We attribute this to the text-based attention weights used for identifying 3D tokens, which effectively enriches the text distribution and implicitly enhances 3D scene understanding while considering the text, thus mitigating overfitting to high-frequency answer distributions.

\subsection{Performance analysis on 3D-RDQA dataset}

\begin{table*}[h]
	\centering
  \addtolength{\tabcolsep}{5pt}
  \setlength{\tabcolsep}{3pt}
 \begin{tabular}{ c | c | c c | c | c }
\toprule
LLM & 3D input & Pre-train & SFT & Strategy of mixture & Accuracy $\uparrow$ \\

\midrule 

\multirow{2}{*}{Qwen2-1.5B} &  & & \cmark & batch concat & 0 \\
& & & \cmark & random sample  & 42.89 \\

\midrule

\multirow{3}{*}{Qwen2-1.5B} & \cmark & \cmark & & batch concat & 0 \\
& \cmark &  & \cmark & batch concat & 58.95 \\
 & \cmark & \cmark & \cmark & batch concat & 76.26 \\

\bottomrule

\end{tabular}
      \caption{\textbf{Further verification on our designed 3D-RDQA dataset.} Two MLP layers are adopt for projector. }
      \label{tab:appendix_final_verification}
\end{table*}

As shown in \cref{tab:appendix_final_verification}, our earlier analysis suggested a potential for the 3D VLMs to memorize answers, and the 0\% accuracy observed in this table reflects this phenomenon. Our 3D-RDQA pair construction involves a Penalty QA item for each question, where the answer is consistently "E," contrasting with the even distribution of Regular QA answers across A, B, C, and D. This design leads to a much higher occurrence of "E" in the training data. As the test set lacks these Penalty QA items, the text-dependent 3D VLM (without the 3D encoder) defaults to the most frequent trained answer, "E," leading to a 0\% accuracy.

It is also crucial to consider the provision of 3D-RDQA pairs. To foster internal adversarial learning within the QA pair, we opt to ensure that each 3D-RDQA pair appeared within the same training batch. This strategy aims to mitigate the learning of spurious correlations that could arise from simple random sampling, although our observations indicate that performance under such sampling remains inferior to that of standard training.

Moreover, the results highlight the benefit of pre-training. However, it is important to note that current 3D-RDQA performance is sensitive to the pre-train dataset. This is because there is a scarcity of large-scale datasets with similar 3D-QA formats. Directly using 3D-LLM Pre for pre-training might lead to suboptimal performance due to discrepancies in 3D-QA format and structure between 3D-LLM Pre and 3D-RDQA. Therefore, we utilize 3D-RDQA itself for pre-train stage here.

\subsection{Performance analysis compared with 3D object-centric methods}
The model designs for 3D scene-centric VLM and 3D object-centric VLM share considerable similarities. A key distinction lies in their approach to feature extraction: 3D scene-centric VLM employs a 3D scene encoder to extract global scene features, subsequently deriving local object proposal features. Conversely, 3D object-centric VLM starts by extracting local object features and then aggregates them to obtain global scene understanding. The commonality of the model architecture suggests that 3D object-centric VLMs may encounter similar limitations.

Recent advancements in 3D object-centric VLM~\cite{huang2023embodied} have demonstrated impressive performance. However, observations from LSceneLLM~\cite{zhi2024lscenellm} indicate a potential bottleneck of them. When the prior knowledge of task-relevant object identities is removed from the recognition model, the performance of LEO~\cite{huang2023embodied} drops to a level comparable to LL3DA, despite LEO utilizing an 8$\times$ larger dataset. This finding aligns with our findings that these models may lack data scaling capabilities on large scale datasets. Furthermore, it implies that a primary advantage of 3D object-centric VLMs stems from the available semantic information associated within defined objects.

\subsection{Performance analysis of 3D-VG}

To facilitate better learning of 3D-VG, we represent each 3D bounding box as $[x,y,z,w,h,l]$, where $x$, $y$ and $z$ denote the coordinate of object center on x-axis, y-axis and z-axis respectively and $w$, $h$ and $l$ denote the width, height and length of the 3D bounding boxes respectively. 
Let $x_{min},y_{min},z_{min}$ represent the minimum value of 3D scene point clouds on x-axis, y-axis and z-axis respectively, and $x_{max},y_{max},z_{max}$ represent the maximum value of 3D scene point clouds on x-axis, y-axis and z-axis respectively.
We normalize the object 3D bounding boxes $[x,y,z,w,h,l]$ based on the input scene:

\begin{equation}
	\label{eq:3DVG_xyz}
	x = \frac{x-x_{min}}{x_{max}-x_{min}}\times g,~
	y = \frac{y-y_{min}}{y_{max}-y_{min}}\times g,~
	z = \frac{z-z_{min}}{z_{max}-z_{min}}\times g,
\end{equation}

where $g$ denotes the maximum value of normalized grid, which is set to 255.

Similarly, we can normalize the 3D bounding box sizes ($w$,$h$,$l$). Considering that the minimum possible value for a 3D bounding box size is zero, we explored two normalization approaches:

\begin{equation}
	\label{eq:3DVG_whl_1}
    \text{\textbf{Signed Normalization:}}~
	w = \frac{w-x_{min}}{x_{max}-x_{min}}\times g,~
	h = \frac{h-y_{min}}{y_{max}-y_{min}}\times g,~
	l = \frac{l-z_{min}}{z_{max}-z_{min}}\times g
\end{equation}

\begin{equation}
	\label{eq:3DVG_whl_2}
    \text{\textbf{Min-zero Normalization:}}~
	w = \frac{w}{x_{max}-x_{min}}\times g,~
	h = \frac{h}{y_{max}-y_{min}}\times g,~
	l = \frac{l}{z_{max}-z_{min}}\times g
\end{equation}

\begin{table*}[h]
	\centering
  \addtolength{\tabcolsep}{5pt}
  \setlength{\tabcolsep}{3pt}
 \begin{tabular}{ c | c | c |l }
\toprule
LLM & Acc@0.25 $\uparrow$ & Dataset & Update \\

 \midrule  
    \multicolumn{3}{l}{\textit{\textbf{Official 3D-LLM}}} \\
    \midrule

flamingo &  21.2 & 675k &\\
BLIP2-opt &  29.6 & 675k &\\
BLIP2-flanT5 & 30.3 & 675k &\\

 \midrule 
    \multicolumn{3}{l}{\textit{\textbf{Signed Normalization}}} \\
    \midrule

\multirow{2}{*}{Qwen2.5-1.5B} & 21.8 & 36k & +3D-VG ScanRefer \\
 & 25.8 & 72k & further +3D-DC ScanRefer \\

 \midrule 
    \multicolumn{3}{l}{\textit{\textbf{Min-zero Normalization}}} \\
    \midrule

\multirow{2}{*}{Qwen2.5-1.5B} & 1.93 & 36k & +3D-VG ScanRefer \\
 & 2.04 & 72k & further +3D-DC ScanRefer \\

\bottomrule

\end{tabular}
      \caption{\textbf{Comparisons to 3D-LLM on 3D-VG.} We train model 1 and 4 epoch for pre-train and SFT stages, respectively.}
      \label{tab:appendix_visual_grounding}
\end{table*}

As shown in \cref{tab:appendix_visual_grounding}, we further conduct in-depth experiments on 3D-VG. Results indicate that the current performance is comparable to that reported in 3D-LLM with Min-zero Normalization, without considering differences in data scale and model architecture. However, when we use Signed Normalization, model demonstrate failing to learn any meaningful 3D-VG knowledge.

Intuitively, Min-zero Normalization should provide more accurate results. However, the near-zero ACC@0.25 indicates a lack of spatial awareness learned from the 3D scene, consistent with our previous observations. Furthermore, while Signed Normalization on 3D bounding box size yields relatively good performance with larger bounding box sizes after normalization, it suggests that the model's performance might stem from encompassing a wider region through box sizes, rather than precise spatial understanding. Overall, our findings suggest that without specific architectural designs, it is challenging for general 3D scene-centric VLMs to learn fine-grained spatial information, leading to inaccurate visual grounding.

\subsection{Expansion to 3D-DC and 3D-VG with relevance discrimination idea}

\begin{figure*}[h]
    \centerline{\includegraphics[width=\textwidth]{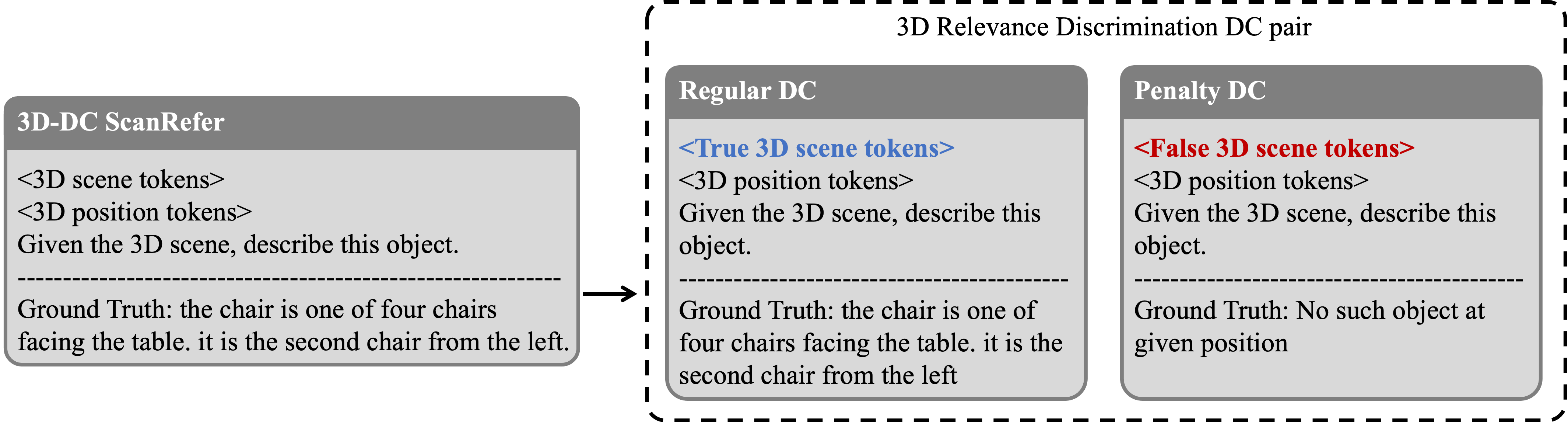}}
    \caption{\textbf{Example visualization of 3D-RDDC (3D Relevance Discrimination Dense Captioning) pair collection.} Unlike in 3D-QA, 3D VLMs on 3D-DC tasks might over-rely on the provided 3D position information rather than the question itself.}
    \vspace{-3mm}
    \label{fig:relevance_discrimination_dc}
\end{figure*}

The core idea of 3D-RDQA is to construct conflicting data pairs regardless of 3D scene features, where a model lacking true 3D spatial understanding would be misled, while a model with genuine 3D vision would not. When applying this concept to 3D-DC, as illustrated in \cref{fig:relevance_discrimination_dc}, the question associated with 3D-DC is often uniform and simple, but the provided 3D position tokens can vary significantly. This variation might lead the model to disregard the 3D scene and the question, instead learning a direct relationship between 3D position tokens and the answer. Therefore, to ensure that information beyond 3D position influences the final answer, and given that we cannot alter the question to avoid the model learning question-answer relationships, we can manipulate the 3D scene tokens. Unlike in \cref{fig:relevance_discrimination_qa}, we cannot directly use 3D tokens from different scenes, as the same 3D position tokens in another scene might hold genuine meaning. Thus, a viable approach is to directly zero out the 3D scene tokens to obtain \textcolor{red}{<False 3D scene tokens>}.

\begin{figure*}[h]
    \centerline{\includegraphics[width=\textwidth]{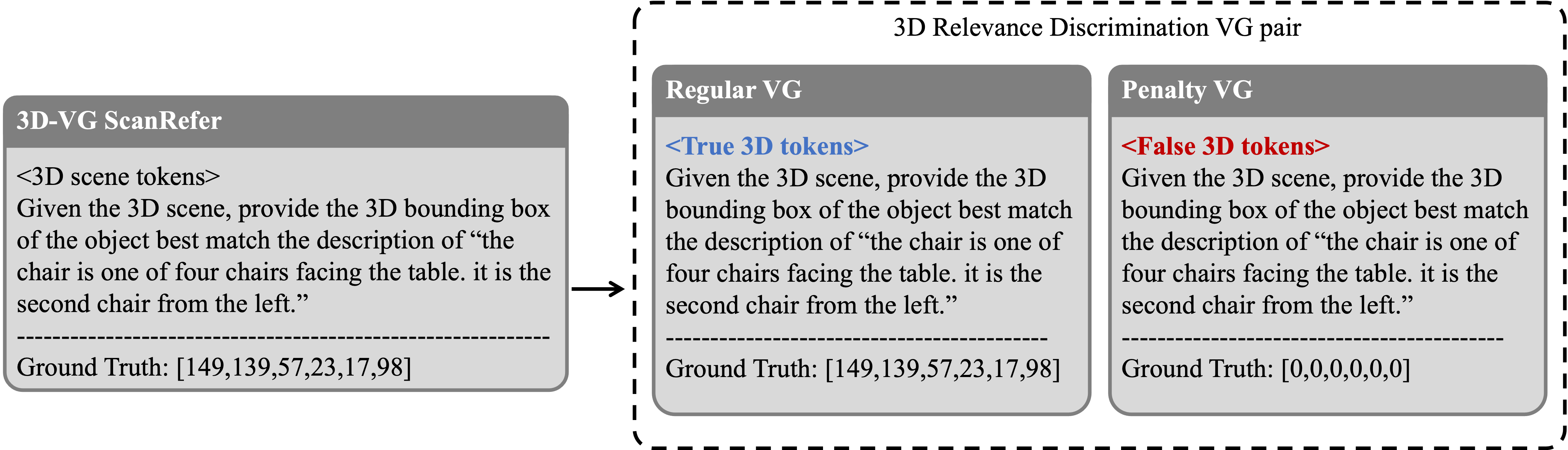}}
    \caption{\textbf{Example visualization of 3D-RDVG (3D Relevance Discrimination Visual Grounding) pair collection.} 3D VLMs on 3D-VG tasks may perform similar to 3D-QA tasks due to the description provided in question, which resulting in similar way to design relevance discrimination data pairs. }
    \vspace{-3mm}
    \label{fig:relevance_discrimination_vg}
\end{figure*}

As illustrated in \cref{fig:relevance_discrimination_vg}, 3D-RDVG employs a similar design pattern to 3D-RDQA. This is motivated by the potential issue in 3D-VG where the rich object descriptions provided in the questions could introduce high question diversity, potentially leading the model to learn a direct mapping between questions and answers. This is analogous to the challenge faced in 3D-QA. Consequently, they can both leverage similar methods to construct relevance discrimination data pairs.

\subsection{Social impact}
The development of robust 3D VLMs holds promise for a wide range of beneficial applications. These include enhanced human-computer interaction in AR/VR environments, improved scene understanding for autonomous navigation in robotics and self-driving vehicles, and more effective training tools in embodied AI simulations. 
However, the technology also presents potential risks for negative societal impacts. For example, the capacity of these models to process and interpret detailed 3D scene information could be misused for surveillance purposes, enabling more sophisticated tracking and monitoring of individuals within private or public spaces.
Consequently, our analysis provides a renewed understanding of 3D VLMs only within the academic context.

%% file: neurips_2025.bbl
\begin{thebibliography}{55}
\providecommand{\natexlab}[1]{#1}
\providecommand{\url}[1]{\texttt{#1}}
\expandafter\ifx\csname urlstyle\endcsname\relax
  \providecommand{\doi}[1]{doi: #1}\else
  \providecommand{\doi}{doi: \begingroup \urlstyle{rm}\Url}\fi

\bibitem[Achlioptas et~al.(2020)Achlioptas, Abdelreheem, Xia, Elhoseiny, and Guibas]{achlioptas2020referit3d}
Panos Achlioptas, Ahmed Abdelreheem, Fei Xia, Mohamed Elhoseiny, and Leonidas Guibas.
\newblock Referit3d: Neural listeners for fine-grained 3d object identification in real-world scenes.
\newblock In \emph{Computer Vision--ECCV 2020: 16th European Conference, Glasgow, UK, August 23--28, 2020, Proceedings, Part I 16}, pages 422--440. Springer, 2020.

\bibitem[Azuma et~al.(2022)Azuma, Miyanishi, Kurita, and Kawanabe]{azuma2022scanqa}
Daichi Azuma, Taiki Miyanishi, Shuhei Kurita, and Motoaki Kawanabe.
\newblock Scanqa: 3d question answering for spatial scene understanding.
\newblock In \emph{proceedings of the IEEE/CVF conference on computer vision and pattern recognition}, pages 19129--19139, 2022.

\bibitem[Chen et~al.(2020)Chen, Chang, and Nie{\ss}ner]{chen2020scanrefer}
Dave~Zhenyu Chen, Angel~X Chang, and Matthias Nie{\ss}ner.
\newblock Scanrefer: 3d object localization in rgb-d scans using natural language.
\newblock In \emph{European conference on computer vision}, pages 202--221. Springer, 2020.

\bibitem[Chen et~al.(2022)Chen, Guhur, Tapaswi, Schmid, and Laptev]{chen2022language}
Shizhe Chen, Pierre-Louis Guhur, Makarand Tapaswi, Cordelia Schmid, and Ivan Laptev.
\newblock Language conditioned spatial relation reasoning for 3d object grounding.
\newblock \emph{Advances in neural information processing systems}, 35:\penalty0 20522--20535, 2022.

\bibitem[Chen et~al.(2023)Chen, Zhu, Chen, Lei, Yu, and Chen]{chen2023end}
Sijin Chen, Hongyuan Zhu, Xin Chen, Yinjie Lei, Gang Yu, and Tao Chen.
\newblock End-to-end 3d dense captioning with vote2cap-detr.
\newblock In \emph{Proceedings of the IEEE/CVF conference on computer vision and pattern recognition}, pages 11124--11133, 2023.

\bibitem[Chen et~al.(2024{\natexlab{a}})Chen, Chen, Zhang, Li, Yu, Fei, Zhu, Fan, and Chen]{chen2024ll3da}
Sijin Chen, Xin Chen, Chi Zhang, Mingsheng Li, Gang Yu, Hao Fei, Hongyuan Zhu, Jiayuan Fan, and Tao Chen.
\newblock Ll3da: Visual interactive instruction tuning for omni-3d understanding reasoning and planning.
\newblock In \emph{Proceedings of the IEEE/CVF Conference on Computer Vision and Pattern Recognition}, pages 26428--26438, 2024{\natexlab{a}}.

\bibitem[Chen et~al.(2024{\natexlab{b}})Chen, Zhu, Li, Chen, Guo, Lei, Yu, Li, and Chen]{chen2024vote2cap}
Sijin Chen, Hongyuan Zhu, Mingsheng Li, Xin Chen, Peng Guo, Yinjie Lei, Gang Yu, Taihao Li, and Tao Chen.
\newblock Vote2cap-detr++: Decoupling localization and describing for end-to-end 3d dense captioning.
\newblock \emph{IEEE Transactions on Pattern Analysis and Machine Intelligence}, 46\penalty0 (11):\penalty0 7331--7347, 2024{\natexlab{b}}.

\bibitem[Chen et~al.(2024{\natexlab{c}})Chen, Yang, Huang, Wang, Xu, Lyu, Lin, and Pang]{chen2024grounded}
Yilun Chen, Shuai Yang, Haifeng Huang, Tai Wang, Runsen Xu, Ruiyuan Lyu, Dahua Lin, and Jiangmiao Pang.
\newblock Grounded 3d-llm with referent tokens.
\newblock \emph{arXiv preprint arXiv:2405.10370}, 2024{\natexlab{c}}.

\bibitem[Chen et~al.(2021)Chen, Gholami, Nie{\ss}ner, and Chang]{chen2021scan2cap}
Zhenyu Chen, Ali Gholami, Matthias Nie{\ss}ner, and Angel~X Chang.
\newblock Scan2cap: Context-aware dense captioning in rgb-d scans.
\newblock In \emph{Proceedings of the IEEE/CVF conference on computer vision and pattern recognition}, pages 3193--3203, 2021.

\bibitem[Chen et~al.(2024{\natexlab{d}})Chen, Wu, Wang, Su, Chen, Xing, Zhong, Zhang, Zhu, Lu, et~al.]{chen2024internvl}
Zhe Chen, Jiannan Wu, Wenhai Wang, Weijie Su, Guo Chen, Sen Xing, Muyan Zhong, Qinglong Zhang, Xizhou Zhu, Lewei Lu, et~al.
\newblock Internvl: Scaling up vision foundation models and aligning for generic visual-linguistic tasks.
\newblock In \emph{Proceedings of the IEEE/CVF conference on computer vision and pattern recognition}, pages 24185--24198, 2024{\natexlab{d}}.

\bibitem[Dai et~al.(2017)Dai, Chang, Savva, Halber, Funkhouser, and Nie{\ss}ner]{dai2017scannet}
Angela Dai, Angel~X Chang, Manolis Savva, Maciej Halber, Thomas Funkhouser, and Matthias Nie{\ss}ner.
\newblock Scannet: Richly-annotated 3d reconstructions of indoor scenes.
\newblock In \emph{Proceedings of the IEEE conference on computer vision and pattern recognition}, pages 5828--5839, 2017.

\bibitem[Dong et~al.(2024)Dong, Zhang, Zang, Cao, Wang, Ouyang, Wei, Zhang, Duan, Cao, et~al.]{dong2024internlm}
Xiaoyi Dong, Pan Zhang, Yuhang Zang, Yuhang Cao, Bin Wang, Linke Ouyang, Xilin Wei, Songyang Zhang, Haodong Duan, Maosong Cao, et~al.
\newblock Internlm-xcomposer2: Mastering free-form text-image composition and comprehension in vision-language large model.
\newblock \emph{arXiv preprint arXiv:2401.16420}, 2024.

\bibitem[Guo et~al.(2023)Guo, Zhang, Zhu, Tang, Ma, Han, Chen, Gao, Li, Li, et~al.]{guo2023point}
Ziyu Guo, Renrui Zhang, Xiangyang Zhu, Yiwen Tang, Xianzheng Ma, Jiaming Han, Kexin Chen, Peng Gao, Xianzhi Li, Hongsheng Li, et~al.
\newblock Point-bind \& point-llm: Aligning point cloud with multi-modality for 3d understanding, generation, and instruction following.
\newblock \emph{arXiv preprint arXiv:2309.00615}, 2023.

\bibitem[Hong et~al.(2023)Hong, Zhen, Chen, Zheng, Du, Chen, and Gan]{hong20233d}
Yining Hong, Haoyu Zhen, Peihao Chen, Shuhong Zheng, Yilun Du, Zhenfang Chen, and Chuang Gan.
\newblock 3d-llm: Injecting the 3d world into large language models.
\newblock \emph{Advances in Neural Information Processing Systems}, 36:\penalty0 20482--20494, 2023.

\bibitem[Huang et~al.(2023{\natexlab{a}})Huang, Wang, Huang, Liu, Cheng, Zhao, Jin, and Zhao]{huang2023chat}
Haifeng Huang, Zehan Wang, Rongjie Huang, Luping Liu, Xize Cheng, Yang Zhao, Tao Jin, and Zhou Zhao.
\newblock Chat-3d v2: Bridging 3d scene and large language models with object identifiers.
\newblock \emph{CoRR}, 2023{\natexlab{a}}.

\bibitem[Huang et~al.(2023{\natexlab{b}})Huang, Yong, Ma, Linghu, Li, Wang, Li, Zhu, Jia, and Huang]{huang2023embodied}
Jiangyong Huang, Silong Yong, Xiaojian Ma, Xiongkun Linghu, Puhao Li, Yan Wang, Qing Li, Song-Chun Zhu, Baoxiong Jia, and Siyuan Huang.
\newblock An embodied generalist agent in 3d world.
\newblock \emph{arXiv preprint arXiv:2311.12871}, 2023{\natexlab{b}}.

\bibitem[Huang et~al.(2023{\natexlab{c}})Huang, Dong, Yang, Huang, Lau, Ouyang, and Zuo]{huang2023clip2point}
Tianyu Huang, Bowen Dong, Yunhan Yang, Xiaoshui Huang, Rynson~WH Lau, Wanli Ouyang, and Wangmeng Zuo.
\newblock Clip2point: Transfer clip to point cloud classification with image-depth pre-training.
\newblock In \emph{Proceedings of the IEEE/CVF International Conference on Computer Vision}, pages 22157--22167, 2023{\natexlab{c}}.

\bibitem[Huang et~al.(2022)Huang, Li, Qu, He, Zuo, and Ouyang]{huang2022frozen}
Xiaoshui Huang, Sheng Li, Wentao Qu, Tong He, Yifan Zuo, and Wanli Ouyang.
\newblock Frozen clip model is efficient point cloud backbone.
\newblock \emph{arXiv preprint arXiv:2212.04098}, 1\penalty0 (6), 2022.

\bibitem[Jiang et~al.(2020)Jiang, Zhao, Shi, Liu, Fu, and Jia]{jiang2020pointgroup}
Li Jiang, Hengshuang Zhao, Shaoshuai Shi, Shu Liu, Chi-Wing Fu, and Jiaya Jia.
\newblock Pointgroup: Dual-set point grouping for 3d instance segmentation.
\newblock In \emph{Proceedings of the IEEE/CVF conference on computer vision and Pattern recognition}, pages 4867--4876, 2020.

\bibitem[Li et~al.(2025)Li, Zhou, Tang, Song, Zeng, Kampffmeyer, Xu, and Liang]{li2025unigs}
Haoyuan Li, Yanpeng Zhou, Tao Tang, Jifei Song, Yihan Zeng, Michael Kampffmeyer, Hang Xu, and Xiaodan Liang.
\newblock Unigs: Unified language-image-3d pretraining with gaussian splatting.
\newblock \emph{arXiv preprint arXiv:2502.17860}, 2025.

\bibitem[Li et~al.(2023)Li, Li, Savarese, and Hoi]{li2023blip}
Junnan Li, Dongxu Li, Silvio Savarese, and Steven Hoi.
\newblock Blip-2: Bootstrapping language-image pre-training with frozen image encoders and large language models.
\newblock In \emph{International conference on machine learning}, pages 19730--19742. PMLR, 2023.

\bibitem[Li et~al.(2024)Li, Zhang, Wang, Ren, Xu, Ma, Liu, and Wei]{li20243dmit}
Zeju Li, Chao Zhang, Xiaoyan Wang, Ruilong Ren, Yifan Xu, Ruifei Ma, Xiangde Liu, and Rong Wei.
\newblock 3dmit: 3d multi-modal instruction tuning for scene understanding.
\newblock In \emph{2024 IEEE International Conference on Multimedia and Expo Workshops (ICMEW)}, pages 1--5. IEEE, 2024.

\bibitem[Lin(2004)]{lin2004rouge}
Chin-Yew Lin.
\newblock Rouge: A package for automatic evaluation of summaries.
\newblock In \emph{Text summarization branches out}, pages 74--81, 2004.

\bibitem[Liu et~al.(2024)Liu, Feng, Xue, Wang, Wu, Lu, Zhao, Deng, Zhang, Ruan, et~al.]{liu2024deepseek}
Aixin Liu, Bei Feng, Bing Xue, Bingxuan Wang, Bochao Wu, Chengda Lu, Chenggang Zhao, Chengqi Deng, Chenyu Zhang, Chong Ruan, et~al.
\newblock Deepseek-v3 technical report.
\newblock \emph{arXiv preprint arXiv:2412.19437}, 2024.

\bibitem[Liu et~al.(2023{\natexlab{a}})Liu, Li, Wu, and Lee]{liu2023visual}
Haotian Liu, Chunyuan Li, Qingyang Wu, and Yong~Jae Lee.
\newblock Visual instruction tuning.
\newblock \emph{Advances in neural information processing systems}, 36:\penalty0 34892--34916, 2023{\natexlab{a}}.

\bibitem[Liu et~al.(2023{\natexlab{b}})Liu, Shi, Kuang, Zhu, Li, Han, Cai, Porikli, and Su]{liu2023openshape}
Minghua Liu, Ruoxi Shi, Kaiming Kuang, Yinhao Zhu, Xuanlin Li, Shizhong Han, Hong Cai, Fatih Porikli, and Hao Su.
\newblock Openshape: Scaling up 3d shape representation towards open-world understanding.
\newblock \emph{Advances in neural information processing systems}, 36:\penalty0 44860--44879, 2023{\natexlab{b}}.

\bibitem[Loshchilov and Hutter(2017)]{loshchilov2017decoupled}
Ilya Loshchilov and Frank Hutter.
\newblock Decoupled weight decay regularization.
\newblock \emph{arXiv preprint arXiv:1711.05101}, 2017.

\bibitem[Ma et~al.(2022)Ma, Yong, Zheng, Li, Liang, Zhu, and Huang]{ma2022sqa3d}
Xiaojian Ma, Silong Yong, Zilong Zheng, Qing Li, Yitao Liang, Song-Chun Zhu, and Siyuan Huang.
\newblock Sqa3d: Situated question answering in 3d scenes.
\newblock \emph{arXiv preprint arXiv:2210.07474}, 2022.

\bibitem[Mishra et~al.(2024)Mishra, Seth, Jain, Pant, Parikh, Jain, and Islam]{mishra2024image}
Swapneel Mishra, Saumya Seth, Shrishti Jain, Vasudev Pant, Jolly Parikh, Rachna Jain, and Sardar~MN Islam.
\newblock Image caption generation using vision transformer and gpt architecture.
\newblock In \emph{2024 2nd International Conference on Advancement in Computation \& Computer Technologies (InCACCT)}, pages 1--6. IEEE, 2024.

\bibitem[Misra et~al.(2021)Misra, Girdhar, and Joulin]{misra2021end}
Ishan Misra, Rohit Girdhar, and Armand Joulin.
\newblock An end-to-end transformer model for 3d object detection.
\newblock In \emph{Proceedings of the IEEE/CVF international conference on computer vision}, pages 2906--2917, 2021.

\bibitem[Papineni et~al.(2002)Papineni, Roukos, Ward, and Zhu]{papineni2002bleu}
Kishore Papineni, Salim Roukos, Todd Ward, and Wei-Jing Zhu.
\newblock Bleu: a method for automatic evaluation of machine translation.
\newblock In \emph{Proceedings of the 40th annual meeting of the Association for Computational Linguistics}, pages 311--318, 2002.

\bibitem[Qi et~al.(2017)Qi, Yi, Su, and Guibas]{qi2017pointnet++}
Charles~Ruizhongtai Qi, Li Yi, Hao Su, and Leonidas~J Guibas.
\newblock Pointnet++: Deep hierarchical feature learning on point sets in a metric space.
\newblock \emph{Advances in neural information processing systems}, 30, 2017.

\bibitem[Qi et~al.(2019)Qi, Litany, He, and Guibas]{qi2019deep}
Charles~R Qi, Or Litany, Kaiming He, and Leonidas~J Guibas.
\newblock Deep hough voting for 3d object detection in point clouds.
\newblock In \emph{proceedings of the IEEE/CVF International Conference on Computer Vision}, pages 9277--9286, 2019.

\bibitem[Tang et~al.(2024)Tang, Han, Li, Yu, Hao, Hu, and Chen]{tang2024minigpt}
Yuan Tang, Xu Han, Xianzhi Li, Qiao Yu, Yixue Hao, Long Hu, and Min Chen.
\newblock Minigpt-3d: Efficiently aligning 3d point clouds with large language models using 2d priors.
\newblock In \emph{Proceedings of the 32nd ACM International Conference on Multimedia}, pages 6617--6626, 2024.

\bibitem[Tang et~al.(2025{\natexlab{a}})Tang, Guo, Wang, Zhang, Chen, Liu, Qu, Wang, Wang, Li, et~al.]{tang2025exploring}
Yiwen Tang, Zoey Guo, Zhuhao Wang, Ray Zhang, Qizhi Chen, Junli Liu, Delin Qu, Zhigang Wang, Dong Wang, Xuelong Li, et~al.
\newblock Exploring the potential of encoder-free architectures in 3d lmms.
\newblock \emph{arXiv preprint arXiv:2502.09620}, 2025{\natexlab{a}}.

\bibitem[Tang et~al.(2025{\natexlab{b}})Tang, Han, Li, Yu, Xu, Hao, Hu, and Chen]{tang2025more}
Yuan Tang, Xu Han, Xianzhi Li, Qiao Yu, Jinfeng Xu, Yixue Hao, Long Hu, and Min Chen.
\newblock More text, less point: Towards 3d data-efficient point-language understanding.
\newblock In \emph{Proceedings of the AAAI Conference on Artificial Intelligence}, pages 7284--7292, 2025{\natexlab{b}}.

\bibitem[Team et~al.(2023)Team, Anil, Borgeaud, Alayrac, Yu, Soricut, Schalkwyk, Dai, Hauth, Millican, et~al.]{team2023gemini}
Gemini Team, Rohan Anil, Sebastian Borgeaud, Jean-Baptiste Alayrac, Jiahui Yu, Radu Soricut, Johan Schalkwyk, Andrew~M Dai, Anja Hauth, Katie Millican, et~al.
\newblock Gemini: a family of highly capable multimodal models.
\newblock \emph{arXiv preprint arXiv:2312.11805}, 2023.

\bibitem[Vedantam et~al.()Vedantam, Zitnick, and Parikh]{vedantamconsensus}
R Vedantam, C~Lawrence Zitnick, and D Parikh.
\newblock Consensus-based image description evaluation.
\newblock In \emph{Proceedings of the IEEE conference on computer vision and pattern recognition}, pages 4566--4575.

\bibitem[Wald et~al.(2019)Wald, Avetisyan, Navab, Tombari, and Nie{\ss}ner]{wald2019rio}
Johanna Wald, Armen Avetisyan, Nassir Navab, Federico Tombari, and Matthias Nie{\ss}ner.
\newblock Rio: 3d object instance re-localization in changing indoor environments.
\newblock In \emph{Proceedings of the IEEE/CVF International Conference on Computer Vision}, pages 7658--7667, 2019.

\bibitem[Wang et~al.(2024)Wang, Bai, Tan, Wang, Fan, Bai, Chen, Liu, Wang, Ge, et~al.]{wang2024qwen2}
Peng Wang, Shuai Bai, Sinan Tan, Shijie Wang, Zhihao Fan, Jinze Bai, Keqin Chen, Xuejing Liu, Jialin Wang, Wenbin Ge, et~al.
\newblock Qwen2-vl: Enhancing vision-language model's perception of the world at any resolution.
\newblock \emph{arXiv preprint arXiv:2409.12191}, 2024.

\bibitem[Wang et~al.(2023)Wang, Huang, Zhao, Zhang, and Zhao]{wang2023chat}
Zehan Wang, Haifeng Huang, Yang Zhao, Ziang Zhang, and Zhou Zhao.
\newblock Chat-3d: Data-efficiently tuning large language model for universal dialogue of 3d scenes.
\newblock \emph{arXiv preprint arXiv:2308.08769}, 2023.

\bibitem[Wu et~al.(2024)Wu, Zhong, Xing, Lai, Liu, Chen, Wang, Zhu, Lu, Lu, et~al.]{wu2024visionllm}
Jiannan Wu, Muyan Zhong, Sen Xing, Zeqiang Lai, Zhaoyang Liu, Zhe Chen, Wenhai Wang, Xizhou Zhu, Lewei Lu, Tong Lu, et~al.
\newblock Visionllm v2: An end-to-end generalist multimodal large language model for hundreds of vision-language tasks.
\newblock \emph{Advances in Neural Information Processing Systems}, 37:\penalty0 69925--69975, 2024.

\bibitem[Xue et~al.(2023)Xue, Gao, Xing, Mart{\'\i}n-Mart{\'\i}n, Wu, Xiong, Xu, Niebles, and Savarese]{xue2023ulip}
Le Xue, Mingfei Gao, Chen Xing, Roberto Mart{\'\i}n-Mart{\'\i}n, Jiajun Wu, Caiming Xiong, Ran Xu, Juan~Carlos Niebles, and Silvio Savarese.
\newblock Ulip: Learning a unified representation of language, images, and point clouds for 3d understanding.
\newblock In \emph{Proceedings of the IEEE/CVF conference on computer vision and pattern recognition}, pages 1179--1189, 2023.

\bibitem[Xue et~al.(2024)Xue, Yu, Zhang, Panagopoulou, Li, Mart{\'\i}n-Mart{\'\i}n, Wu, Xiong, Xu, Niebles, et~al.]{xue2024ulip}
Le Xue, Ning Yu, Shu Zhang, Artemis Panagopoulou, Junnan Li, Roberto Mart{\'\i}n-Mart{\'\i}n, Jiajun Wu, Caiming Xiong, Ran Xu, Juan~Carlos Niebles, et~al.
\newblock Ulip-2: Towards scalable multimodal pre-training for 3d understanding.
\newblock In \emph{Proceedings of the IEEE/CVF Conference on Computer Vision and Pattern Recognition}, pages 27091--27101, 2024.

\bibitem[Yang et~al.(2024)Yang, Yang, Hui, Zheng, Yu, Zhou, Li, Li, Liu, Huang, Dong, Wei, Lin, Tang, Wang, Yang, Tu, Zhang, Ma, Yang, Xu, Zhou, Bai, He, Lin, Dang, Lu, Chen, Yang, Li, Xue, Ni, Zhang, Wang, Peng, Men, Gao, Lin, Wang, Bai, Tan, Zhu, Li, Liu, Ge, Deng, Zhou, Ren, Zhang, Wei, Ren, Liu, Fan, Yao, Zhang, Wan, Chu, Liu, Cui, Zhang, Guo, and Fan]{yang2024qwen2}
An Yang, Baosong Yang, Binyuan Hui, Bo Zheng, Bowen Yu, Chang Zhou, Chengpeng Li, Chengyuan Li, Dayiheng Liu, Fei Huang, Guanting Dong, Haoran Wei, Huan Lin, Jialong Tang, Jialin Wang, Jian Yang, Jianhong Tu, Jianwei Zhang, Jianxin Ma, Jianxin Yang, Jin Xu, Jingren Zhou, Jinze Bai, Jinzheng He, Junyang Lin, Kai Dang, Keming Lu, Keqin Chen, Kexin Yang, Mei Li, Mingfeng Xue, Na Ni, Pei Zhang, Peng Wang, Ru Peng, Rui Men, Ruize Gao, Runji Lin, Shijie Wang, Shuai Bai, Sinan Tan, Tianhang Zhu, Tianhao Li, Tianyu Liu, Wenbin Ge, Xiaodong Deng, Xiaohuan Zhou, Xingzhang Ren, Xinyu Zhang, Xipin Wei, Xuancheng Ren, Xuejing Liu, Yang Fan, Yang Yao, Yichang Zhang, Yu Wan, Yunfei Chu, Yuqiong Liu, Zeyu Cui, Zhenru Zhang, Zhifang Guo, and Zhihao Fan.
\newblock Qwen2 technical report, 2024.

\bibitem[Yu et~al.(2022)Yu, Tang, Rao, Huang, Zhou, and Lu]{yu2022point}
Xumin Yu, Lulu Tang, Yongming Rao, Tiejun Huang, Jie Zhou, and Jiwen Lu.
\newblock Point-bert: Pre-training 3d point cloud transformers with masked point modeling.
\newblock In \emph{Proceedings of the IEEE/CVF conference on computer vision and pattern recognition}, pages 19313--19322, 2022.

\bibitem[Zeng et~al.(2023)Zeng, Jiang, Mao, Han, Ye, Huang, Yeung, Yang, Liang, and Xu]{zeng2023clip2}
Yihan Zeng, Chenhan Jiang, Jiageng Mao, Jianhua Han, Chaoqiang Ye, Qingqiu Huang, Dit-Yan Yeung, Zhen Yang, Xiaodan Liang, and Hang Xu.
\newblock Clip2: Contrastive language-image-point pretraining from real-world point cloud data.
\newblock In \emph{Proceedings of the IEEE/CVF conference on computer vision and pattern recognition}, pages 15244--15253, 2023.

\bibitem[Zhang et~al.(2022)Zhang, Guo, Zhang, Li, Miao, Cui, Qiao, Gao, and Li]{zhang2022pointclip}
Renrui Zhang, Ziyu Guo, Wei Zhang, Kunchang Li, Xupeng Miao, Bin Cui, Yu Qiao, Peng Gao, and Hongsheng Li.
\newblock Pointclip: Point cloud understanding by clip.
\newblock In \emph{Proceedings of the IEEE/CVF conference on computer vision and pattern recognition}, pages 8552--8562, 2022.

\bibitem[Zhang et~al.(2023{\natexlab{a}})Zhang, Wang, Qiao, Gao, and Li]{zhang2023learning}
Renrui Zhang, Liuhui Wang, Yu Qiao, Peng Gao, and Hongsheng Li.
\newblock Learning 3d representations from 2d pre-trained models via image-to-point masked autoencoders.
\newblock In \emph{Proceedings of the IEEE/CVF Conference on Computer Vision and Pattern Recognition}, pages 21769--21780, 2023{\natexlab{a}}.

\bibitem[Zhang et~al.(2023{\natexlab{b}})Zhang, Gong, and Chang]{zhang2023multi3drefer}
Yiming Zhang, ZeMing Gong, and Angel~X Chang.
\newblock Multi3drefer: Grounding text description to multiple 3d objects.
\newblock In \emph{Proceedings of the IEEE/CVF International Conference on Computer Vision}, pages 15225--15236, 2023{\natexlab{b}}.

\bibitem[Zhi et~al.(2024)Zhi, Chen, Li, Ma, Sun, Xiang, Lei, Tan, and Gan]{zhi2024lscenellm}
Hongyan Zhi, Peihao Chen, Junyan Li, Shuailei Ma, Xinyu Sun, Tianhang Xiang, Yinjie Lei, Mingkui Tan, and Chuang Gan.
\newblock Lscenellm: Enhancing large 3d scene understanding using adaptive visual preferences.
\newblock \emph{arXiv preprint arXiv:2412.01292}, 2024.

\bibitem[Zhou et~al.(2023)Zhou, Wang, Ma, Liu, Huang, and Wang]{Uni3d}
Junsheng Zhou, Jinsheng Wang, Baorui Ma, Yu-Shen Liu, Tiejun Huang, and Xinlong Wang.
\newblock Uni3d: Exploring unified 3d representation at scale.
\newblock \emph{arXiv preprint arXiv:2310.06773}, 2023.

\bibitem[Zhu et~al.(2024)Zhu, Wang, Zhang, Pang, and Liu]{zhu2024llava}
Chenming Zhu, Tai Wang, Wenwei Zhang, Jiangmiao Pang, and Xihui Liu.
\newblock Llava-3d: A simple yet effective pathway to empowering lmms with 3d-awareness.
\newblock \emph{arXiv preprint arXiv:2409.18125}, 2024.

\bibitem[Zhu et~al.(2023{\natexlab{a}})Zhu, Chen, Shen, Li, and Elhoseiny]{zhu2023minigpt}
Deyao Zhu, Jun Chen, Xiaoqian Shen, Xiang Li, and Mohamed Elhoseiny.
\newblock Minigpt-4: Enhancing vision-language understanding with advanced large language models.
\newblock \emph{arXiv preprint arXiv:2304.10592}, 2023{\natexlab{a}}.

\bibitem[Zhu et~al.(2023{\natexlab{b}})Zhu, Zhang, He, Guo, Zeng, Qin, Zhang, and Gao]{zhu2023pointclip}
Xiangyang Zhu, Renrui Zhang, Bowei He, Ziyu Guo, Ziyao Zeng, Zipeng Qin, Shanghang Zhang, and Peng Gao.
\newblock Pointclip v2: Prompting clip and gpt for powerful 3d open-world learning.
\newblock In \emph{Proceedings of the IEEE/CVF international conference on computer vision}, pages 2639--2650, 2023{\natexlab{b}}.

\end{thebibliography}
